\documentclass[runningheads]{llncs}

\usepackage{eccv}

\usepackage{eccvabbrv}

\usepackage[utf8]{inputenc} 
\usepackage[T1]{fontenc}    
\usepackage{url}            
\usepackage{booktabs}       
\usepackage{amsfonts}       
\usepackage{nicefrac}       
\usepackage{microtype}      
\usepackage{graphicx}
\usepackage{subcaption}
\usepackage{xspace}
\usepackage{mdframed}
\usepackage{enumitem}
\usepackage{bm}
\usepackage{pifont} 
\usepackage{amssymb}
\usepackage{wrapfig}

\usepackage{algorithm}
\usepackage{algorithmicx}
\usepackage{algpseudocode}
\usepackage{multirow}
\usepackage{algpseudocode}
\usepackage{amsmath}
\usepackage{xcolor}
\usepackage{tcolorbox}
\usepackage{listings}
\usepackage{xcolor}
\usepackage{makecell}  
\usepackage{soul}
\usepackage{bbding}
\usepackage{tabularx}
\usepackage{array}
\usepackage[table]{xcolor}

\usepackage{siunitx}
\definecolor{azure}{rgb}{0.0, 0.5, 1.0}
\usepackage{colortbl}
\usepackage[table]{xcolor}
\usepackage{nicematrix}

\newcommand{\ours}{\mbox{{DART}}\xspace}
\newcommand{\oursFull}{{{\textbf{D}omain \textbf{AR}i\textbf{T}hmetic}}\xspace}

\usepackage[accsupp]{axessibility}  

\usepackage{hyperref}

\usepackage{orcidlink}

\usepackage{xspace}

\newcommand{\piOfive}{\ensuremath{\pi_{0.5}}\xspace}
\newcommand{\pifast}{\ensuremath{\pi_{0}\text{-FAST}}\xspace}

\begin{document}

\title{Domain Arithmetic: One-Shot VLA Adaptation under Environmental Shifts} 

\author{Taewook Kang\thanks{These authors contributed equally.}\orcidlink{0009-0005-2669-6135} \and
Taeheon Kim\protect\footnotemark[1]\orcidlink{0009-0004-7642-849X} \and
Donghyun Shin\orcidlink{0009-0009-2160-6814} \and
Jonghyun Choi\textsuperscript{$\dagger$}\orcidlink{0000-0002-7934-8434}}

\authorrunning{T.~Kang et al.}

\institute{Seoul National University\\
\email{\{tw.kang, thkim0305, dawnme, jonghyunchoi\}@snu.ac.kr}}

\maketitle

\begingroup
\renewcommand{\thefootnote}{$\dagger$}
\footnotetext{JC is with ECE, IPAI and ASRI in SNU and a corresponding author.}
\endgroup

\begin{abstract}
Vision-Language-Action (VLA) models often fail to perform the same learned tasks under environmental shifts, such as changes in camera pose and shifts to a different but similar robot (\eg, from Panda to UR5e).
Adapting these models to the shifted environment (\ie, target domain) often requires training on multiple demonstrations for each task, which are costly to collect.
To reduce the burden of data curation and training, we propose an analogy-based method that adapts VLA models under environmental shifts through weight vector arithmetic with domain-specific information addition, named \oursFull (\textbf{\ours}).
Unlike prior approaches, \ours requires collecting only \emph{a single demonstration}, enabling efficient adaptation.
To accurately isolate domain-specific information for addition, \ours performs subspace alignment between singular components in weight vectors to filter out noisy components.
In both simulated and real-world experiments, \ours outperforms existing VLA adaptation methods in one-shot scenarios across diverse visual and embodiment shifts. 
Code is available at {\color{magenta} \url{https://github.com/snumprlab/dart}}.
    
\end{abstract}

\section{Introduction}
\label{sec:intro}

Vision-Language-Action (VLA) models trained on large-scale corpora show strong multi-task capabilities~\cite{zitkovich2023rt2, kim2024openvla, kim2025oft, black2024pi_0, zhou2025pi05, qu2025eo1, bjorck2025gr00t}. 
Despite their success within trained environments, \ie, \emph{source domain}, VLA models face challenges when deployed in new environments to perform learned tasks, a common real-world deployment scenario.
These environmental shifts involve altered camera poses, distinct sensor calibrations, or embodiment modifications, leading to substantial performance degradation~\cite{xie2024decomposing, li2024evaluating, gao2025taxonomy, zhang2025effective, zhu2025efficient, fei2025liberoplus, zhou2025liberopro}.
Thus, post-hoc adaptation remains essential to guarantee reliable execution in the shifted environment, \ie, \emph{target domain}.
However, existing VLA adaptation approaches~\cite{li2025fla, dey2025revla,fei2025liberoplus, yadav2025retain,abouzeid2025geoaware, wilcox2025adapt3r} often require extensive expert demonstrations for \emph{every} policy task in the target domain, resulting in severe deployment bottlenecks~\cite{walke2023bridgedata,mandlekar2023mimicgen,yu2023rosie}. 
Also, fine-tuning on limited data often fails to generalize to unseen tasks~\cite{dey2025revla, yadav2025retain}.

\begin{figure}[tb]
  \centering
  \includegraphics[width=\linewidth]{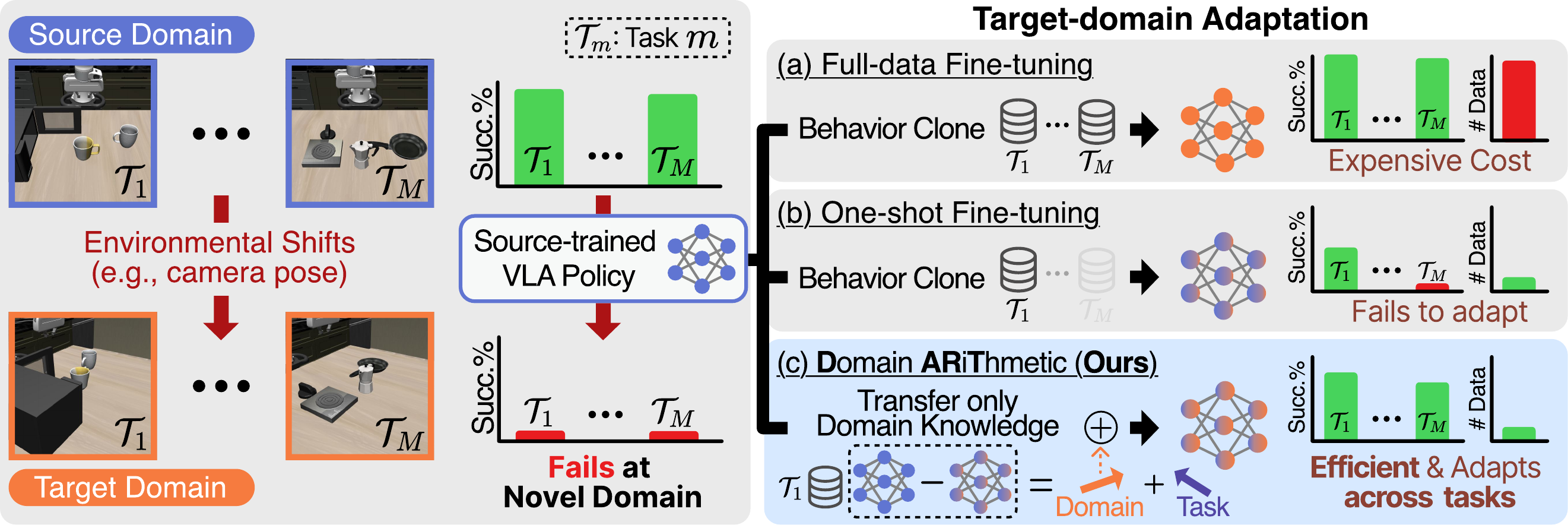}
  \caption{\textbf{One-shot VLA adaptation under environmental shifts.}
  Environmental shifts can cause a source-trained VLA policy to fail in the novel target domain. 
  (a) Full-data fine-tuning adapts successfully but requires task-wise demonstrations, incurring expensive data collection costs. 
  (b) One-shot fine-tuning is data-light but often fails to adapt across tasks. 
  (c) Our one-shot adaptation extracts domain-specific directions from fine-tuned weights and adapts the policy via weight arithmetic.
  }
  \label{fig:teaser}
\end{figure}

For practical policy deployment, extreme data efficiency for adaptation is essential in settings where collecting task-wise demonstrations at scale is typically infeasible, such as household environments.
Thus, we aim for \textbf{one-shot VLA adaptation} where a policy adapts under environmental shifts using only a \emph{single} demonstration of a \emph{single} task.
To enable this, we leverage the insight that a single demonstration can provide transferable domain knowledge.
It allows a source-trained base VLA model to harness its learned task capabilities to solve the same tasks in the target domain, without relearning from scratch (\cref{fig:teaser}).
 
To substantiate this idea, we analyze why one-shot fine-tuned models fail to adapt.
Through subspace alignment analysis, we find that their parameter changes from a base model, \ie, update-vectors, are predominantly task-specific, with small domain-specific directions present.
From this, we further study how these directions coexist.
We conjecture that the task and domain directions are additively composable, inspired by the disentangled weights for each task in vision-language models~\cite{ilharco2023TA, ortiz2023tangent, yun2025soma},
and we find consistent results supporting this structural property in a VLA model.

Building on this observation, we propose \oursFull (\textbf{\ours}), an analogy-based framework for VLA adaptation inspired by weight arithmetic~\cite{ilharco2023TA, zhao2025adamergex,thakkar2024mergealign}.
Similar to \texttt{``queen = king + woman - man''}, we add a \emph{domain vector} that captures the environmental shift to the base model, transferring its multi-task capabilities to the target domain without additional data or architectural changes.
To isolate the domain vector, we subtract the source-domain update-vector from the target-domain update-vector to cancel out shared task-specific directions, where each update-vector is fine-tuned on one-shot data of the same task.

However, direct subtraction can retain source-domain artifacts and amplify fine-tuning noise~\cite{yadav2023ties, yu2024dare, yang2025resm}, corrupting the extracted domain vector.
To address this, motivated by the low-rank structure of update-vectors and subspace aligning property for relevant features in model merging~\cite{marczak2025isoc, gargiulo2025tsv, stoica2025knots},
we introduce \textbf{subspace filtering}, which filters misaligned subspace basis vectors between source and target updates for subtraction,  and \textbf{subspace scaling}, which down-weights noisy domain vectors based on source-target subspace alignment. 

In simulated and real-world experiments with \piOfive~\cite{zhou2025pi05} and \pifast~\cite{pertsch2025fast}, \ours outperforms existing VLA adaptation baselines under one-shot scenarios across diverse visual and embodiment shifts. Moreover, our approach enables fast, hyperparameter-robust adaptation and merging across multiple target domains.

Our \textbf{contributions} are as follows: (i) Empirical evidence shows that one-shot fine-tuned weights decompose into shareable, additive task- and domain-specific directions. (ii) From this observation, we propose \ours that extracts a reusable domain vector from one-shot fine-tuned weights by removing task-specific directions through an analogy operation. (iii) Our approach outperforms prior VLA adaptation methods across diverse simulated and real-world experiments.

\section{Related Work}
\label{sec:related_work}

\subsubsection{Domain adaptation for Vision-Language-Action models.}
By integrating pretrained vision-language backbones~\cite{liu2023llava,beyer2024paligemma} with large-scale robotic datasets~\cite{o2024oxe,walke2023bridgedata,khazatsky2024droid}, VLA models such as RT-2~\cite{zitkovich2023rt2}, OpenVLA~\cite{kim2024openvla, kim2025oft}, and the $\pi$ series~\cite{black2024pi_0, pertsch2025fast, zhou2025pi05, intelligence2025pi06} have shown strong performance in a wide range of tasks.
However, they often require adaptation in novel environments to avoid performance degradation~\cite{xie2024decomposing, li2024evaluating, fei2025liberoplus, zhou2025liberopro, wilcox2025adapt3r}.
A common VLA adaptation approach trains with diverse augmentations to improve generalization~\cite{fei2025liberoplus, chen2024rovi, yang2025novel, li2025fla}, but it relies on fine-tuning with additional training data, incurring prohibitive data collection costs.
Some approaches reduce this data burden by leveraging semantic-rich visual features~\cite{nair2022r3m, dey2025revla, lin2025evo-0} or introduce architectural modifications~\cite{abouzeid2025geoaware, wilcox2025adapt3r, fu2025mergevla}. Yet these choices limit generalizability across backbones and deployment setups.
Test-time adaptation methods~\cite{choi2026scale,liu2026vls} adapt at inference time without extra VLA training, but are mainly targeting limited shift regimes.
To address these limitations, we propose an architecture-agnostic VLA adaptation method with minimal target-domain data under diverse shifts.

\subsubsection{Analogy operation using weight arithmetic.}
Task Arithmetic (TA)~\cite{ilharco2023TA} manipulates models using weight-space arithmetic operations, primarily through \emph{merging} (\ie, addition) to compose capabilities, and \emph{analogy} (\ie, \texttt{``queen - king = woman - man''}) to estimate parameter changes that transfer target properties.
While merging has advanced through interference mitigation~\cite{yadav2023ties,sun2025cat,cheng2025wudi,yang2025resm} and subspace alignment~\cite{marczak2025isoc,gargiulo2025tsv,Wei2026optmerge,panariello2025core,stoica2025knots}, analogy remains limited to direct subtraction, applied in language models for cross-lingual adaptation~\cite{chronopoulou2024language,zhao2025adamergex} and human-alignment transfer~\cite{huang2024chatvector,thakkar2024mergealign}.
In VLA models, existing work focuses exclusively on merging to improve generalization~\cite{dey2025revla,yadav2025retain,sima2026kai0}, or compose skills~\cite{fu2025mergevla,wang2023robotfleet,lawson2024mergingdt}.
However, merging cannot selectively transfer specific capabilities such as domain knowledge while preserving others.
This limitation motivates revisiting analogy for efficient VLA adaptation. 
Our empirical analysis reveals disentangled, additive task- and domain-specific directions in one-shot fine-tuned VLA models, making analogy a natural fit for isolating domain vector.

\section{Preliminaries}

\subsubsection{VLA fine-tuning.}

Let $\pi_{\theta}(\mathbf{a}_t | \mathbf{o}_t, \mathcal{T})$ denote a Vision-Language-Action (VLA) policy parameterized by $\theta$, which maps an observation $\mathbf{o}_t$ (\eg, third-person and wrist camera images) and a task instruction $\mathcal{T}$ (\eg, language prompts) to a distribution over actions $\mathbf{a}_t$ at time step $t$. 
Let $\mathcal{E}_{\text{src}}$ and $\mathcal{E}_{\text{tgt}}$ represent the source and target domains, respectively, where $\mathcal{E}_{\text{tgt}}$ introduces environmental shifts (\eg, camera viewpoint or embodiment changes) in a single (or small number of) environment of $\mathcal{E}_{\text{src}}$. 
We assume access to base policy $\theta_{0}$ that has been trained to solve a suite of policy tasks $\bm{\mathcal{T}} = \{\mathcal{T}_1, \dots, \mathcal{T}_M\}$ within $\mathcal{E}_{\text{src}}$. 
For the target domain $\mathcal{E}_\text{tgt}$, we are given a dataset $\mathcal{D}_{m,\text{tgt}} = \{ (\mathbf{o}_{t}^{\text{tgt}}, \mathbf{a}_{t}, \mathcal{T}_m) \}$ comprising a \emph{single} demonstration for one \emph{adaptation} task $\mathcal{T}_m \in \bm{\mathcal{T}}$ collected per environment in $\mathcal{E}_\text{tgt}$. 

Our objective is to produce adapted parameters $\theta^{*}$ such that $\pi_{\theta^{*}}$ performs well in $\mathcal{E}_{\text{tgt}}$ across \emph{all tasks} in $\bm{\mathcal{T}}$, despite observing only a one-shot, single-task supervision $\mathcal{D}_{m,\text{tgt}}$ per environment. 
For adapting VLA models, we consider the adaptation through behavior cloning (BC) fine-tuning~\cite{zitkovich2023rt2, kim2024openvla, black2024pi_0}. 
Initializing with $\theta_0$, we obtain the target-domain fine-tuned parameters $\theta_{m,\text{tgt}}$ by minimizing a BC objective over the actions in $\mathcal{D}_{m,\text{tgt}}$. 
We then evaluate $\theta_{m,\text{tgt}}$ within $\mathcal{E}_\text{tgt}$ across all tasks in $\bm{\mathcal{T}}$, including $\mathcal{T}_m$ and remaining \emph{held-out} tasks $\mathcal{T}_{k\neq m} \in \bm{\mathcal{T}}$.

\subsubsection{Update-vector.}
To understand the property of fine-tuned weights, we analyze how the model changes through optimization.
Building upon Task Arithmetic~\cite{ilharco2023TA}, we represent an adaptation as a parameter change.
Let $\theta^{(l)}_0$ and $\theta^{(l)}_{m,\text{tgt}} \in \mathbb{R}^{d_{\text{out}} \times d_{\text{in}}}$ denote the weights of layer $l$ in the base and fine-tuned models, respectively.
We define the target-domain \emph{update-vector} in layer $l$, $\mathrm\Delta^{(l)}_{m,\text{tgt}}$, as:
\begin{equation}
\label{eq:update_vector}
\mathrm{\Delta}^{(l)}_{m,\text{tgt}} = \theta^{(l)}_{m,\text{tgt}} - \theta^{(l)}_0,
\end{equation}
and denote the full update-vector by $\mathrm\Delta_{m,\text{tgt}} = \{\mathrm\Delta^{(l)}_{m,\text{tgt}}\}_{l=1}^{L}$.

\vspace{-0.5em}
\section{Analysis of One-shot Fine-tuning Failures}
\label{sec:analysis}

\begin{figure}[tb]
  \centering
  \begin{subfigure}{0.46\linewidth}
    \includegraphics[width=\linewidth]{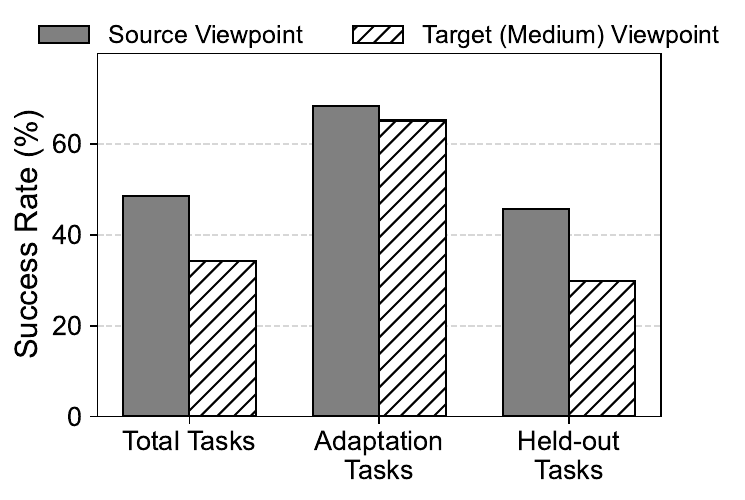}
    \caption{\textbf{One-shot fine-tuning performance.}}
    \label{fig:ft_medium}
  \end{subfigure}
  \hfill
  \begin{subfigure}{0.52\linewidth}
    \includegraphics[width=0.98\linewidth]{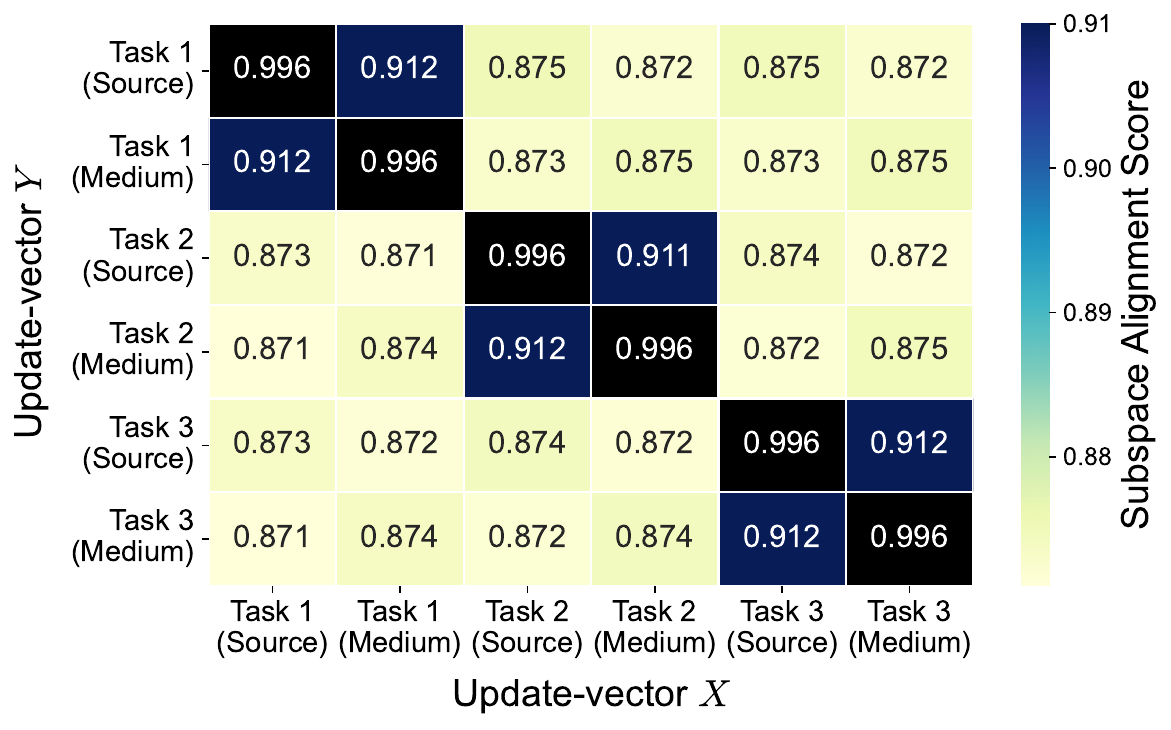}
    \caption{\textbf{Alignment $\gamma(X,Y)$ between update-vectors.}}
    \label{fig:heatmap_task}
  \end{subfigure}
  \vspace{-0.25em}
   \caption{\textbf{Properties of one-shot fine-tuning.}
  (a) The model is fine-tuned on adaptation tasks in target (\texttt{Medium}) camera viewpoint. Performance remains high in adaptation tasks but generalizes poorly to other held-out tasks.
  (b) Subspace alignment $\gamma(\cdot,\cdot)$ among update-vectors $\mathrm\Delta_{m,\text{tgt}} = \theta_{m,\text{tgt}} - \theta_{0}$ on $m\in\{1,2,3\}$ and $\text{tgt}\in\{\text{\texttt{Source}, \texttt{Medium}}\}$. 
  Vectors align for the same task and domain, showing task- and domain-shared directions.
  }
  \label{fig:few_shot_ft}
\end{figure}

Given the substantial cost of collecting data across tasks in each new environment, we consider \textbf{one-shot adaptation} using a single demonstration $\mathcal{D}_{m,\text{tgt}}$ for one adaptation task $\mathcal{T}_m$ in the target domain $\mathcal{E}_\text{tgt}$.
We conjecture in $\S$~\ref{sec:intro} that $\mathcal{D}_{m,\text{tgt}}$ can provide transferable domain knowledge, enabling the base model $\theta_{0}$ to harness its multi-task capabilities in $\mathcal{E}_{\text{tgt}}$.
A natural approach is to fine-tune $\theta_{0}$ on $\mathcal{D}_{m,\text{tgt}}$, yielding $\theta_{m,\text{tgt}}$.
However, on the LIBERO~\cite{liu2023libero} benchmark with \piOfive~\cite{zhou2025pi05}, \cref{fig:ft_medium} shows that one-shot fine-tuning generally fails on \emph{held-out} tasks $\mathcal{T}_{k\neq m}$ when evaluated in $\mathcal{E}_{\text{tgt}}$, where $\text{tgt} = \texttt{Medium}$ (viewpoint shift relative to the source domain, see $\S$~\ref{sec:experiments} for details).
At the same time, performance of $\theta_{m,\text{tgt}}$ on $\mathcal{T}_m$ remains higher than that on $\mathcal{T}_{k\neq m}$ when evaluated in the source domain (\ie, \texttt{Source} viewpoint).
This indicates that one-shot model updates primarily capture task-specific behavior rather than adapting to the target domain across tasks.

\subsection{Subspace Alignment Between Update-Vectors}
\label{sec:similarity}

To understand why one-shot fine-tuned weights fail in multi-task settings, we inspect the components of parameter updates.
In particular, we analyze the similarity between layer-wise update-vectors $\mathrm\Delta^{(l)}_{m,\text{tgt}}$ across tasks and domains.
Since models encoding the same knowledge or capability exhibit high similarity~\cite{jang2024modelstock, yadav2023ties}, we expect to see the same pattern for update-vectors trained on the same task or domain.
As subspace similarity is a strong predictor of transferability and downstream performance~\cite{chen2019bss, stoica2025knots, gargiulo2025tsv, marczak2025isoc}, we quantify the similarity between two update-vectors $\mathrm\Delta^{(l)}_i$ and $\mathrm\Delta^{(l)}_j$ at layer $l$ using the subspace alignment score~\cite{marczak2025isoc}:
\begin{equation}
\label{eq:sar}
\gamma^{(l)}(\mathrm\Delta_i,\mathrm\Delta_j)
=
\frac{\left\lVert U^{(l)}_j {U^{(l)\top}_j} \mathrm\Delta^{(l)}_i \right\rVert_F}{\left\lVert \mathrm\Delta^{(l)}_i \right\rVert_F},
\end{equation}
where $U^{(l)}_j$ are left singular vectors from the Singular Value Decomposition (SVD):
$\mathrm\Delta^{(l)}_j = U^{(l)}_j \mathrm{\Sigma}^{(l)}_j {V^{(l)}_j}^{\top}$.\footnote{We use top-$r$ vectors of $U^{(l)}_j$ with $r = \min \left\{ r':\frac{\sum_{i=r'+1}^R \sigma_i^2}{\sum_{i=1}^R \sigma_i^2} \le 0.05^2 \right\}$, following \cite{marczak2025isoc}.}
Conceptually, $\gamma^{(l)}(\mathrm\Delta_i,\mathrm\Delta_j)$ measures the fraction of $\mathrm\Delta^{(l)}_i$ that can be represented by the subspace of $\mathrm\Delta^{(l)}_j$.
In practice, we aggregate $\gamma^{(l)}(\cdot,\cdot)$ across layers into $\gamma(\cdot,\cdot) = \frac{1}{L} \sum_{l=1}^L \gamma^{(l)}(\cdot,\cdot)$ to obtain a single alignment score.\footnote{We exclude layers with one-dimensional (\eg, biases and normalization) weights.}

As shown in \cref{fig:heatmap_task}, update-vectors from the same adaptation task exhibit strong alignment across domains, indicating that one-shot updates are dominated by task-specific directions.
At the same time, we observe slightly higher overlap among vectors targeting the same domain across different tasks than those targeting different domains, suggesting a consistent domain-shared component in the updates.
This indicates the presence of domain knowledge within the weight space that can be learned through one-shot fine-tuning.

\subsection{Additive Composition of One-shot Update-Vector}
\label{sec:hypothesis}

\begin{figure}[tb]
  \centering
  \begin{subfigure}{0.45\linewidth}
    \includegraphics[width=0.99\linewidth]{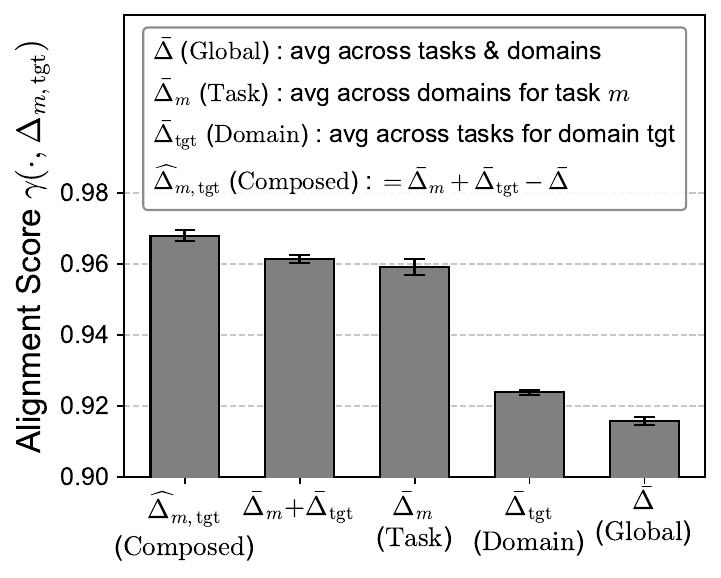}
    \caption{\textbf{Average alignment $\gamma(\cdot,\mathrm\Delta_{m,\text{tgt}})$ with prototypes and their composition.}}
    \label{fig:additiv_test}
  \end{subfigure}
  \hfill
  \begin{subfigure}{0.53\linewidth}
    \includegraphics[width=0.96\linewidth]{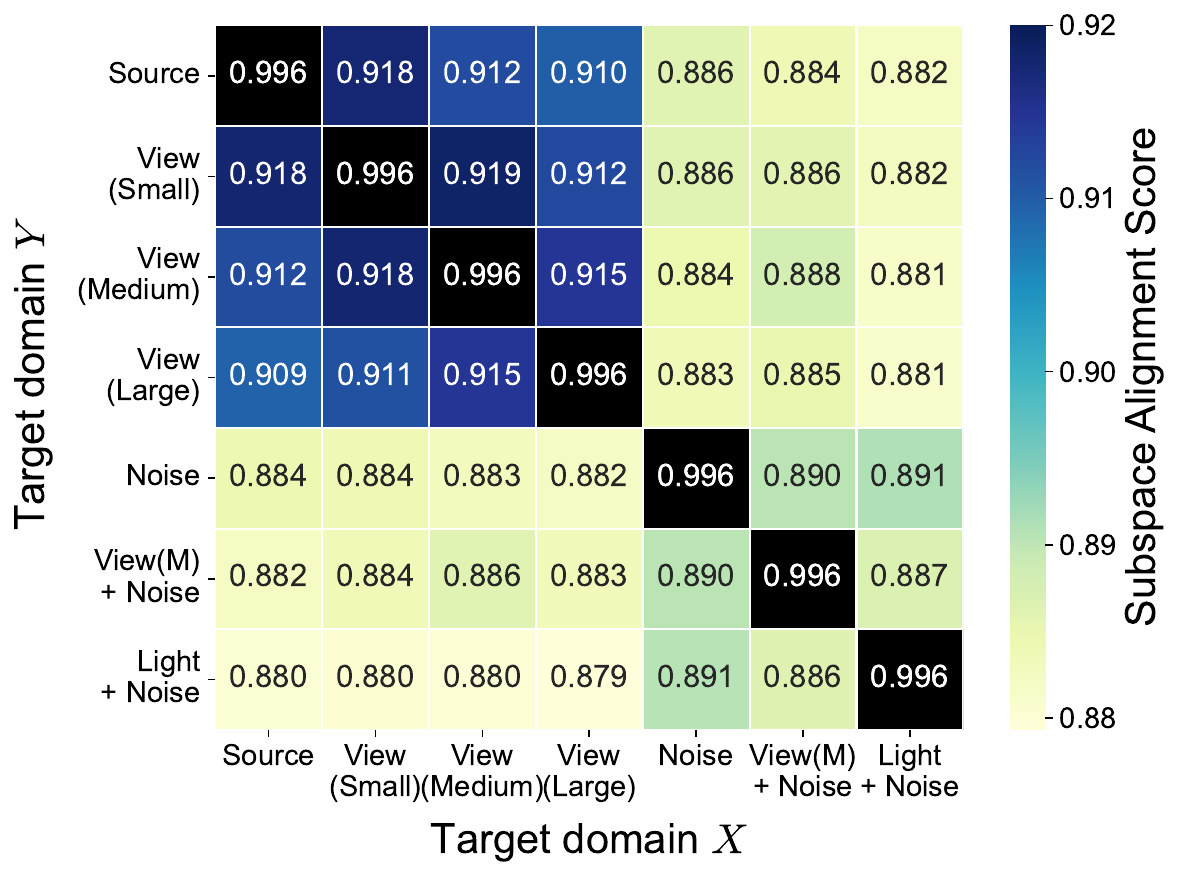}
    \caption{\textbf{Alignment $\gamma(X,Y)$ across target domains.}
    }
    \label{fig:heatmap_domain}
  \end{subfigure}
  \vspace{-0.5em}
  \caption{\textbf{Additive task-domain directions in update-vectors.}
  (a) Prototypes are computed by 16 update-vectors from 4 tasks and 4 domains.
  Strong alignment by additive composition suggests orthogonal, linearly combinable task and domain components.
  (b) \texttt{View} is viewpoint shift, \texttt{Noise} is camera noise, \texttt{Light} is light change.
  Alignment among similar domain shifts shows that domain components are structured and correlated with semantic of domain shifts.
  }
  \vspace{-0.5em}
  \label{fig:subspace_align_score}
\end{figure}

Building on observation of task and domain-shared directions in update-vectors in $\S$~\ref{sec:similarity}, we investigate how these directions coexist. 
Prior work shows that fine-tuning vision and language models for different tasks often updates orthogonal~\cite{yun2025soma}, disentangled weight subspaces~\cite{ilharco2023TA, ortiz2023tangent, jin2025fine}, enabling interference-free additive composition of task capabilities. 
Because VLA models are built upon these pretrained vision-language backbones~\cite{zhou2025pi05, kim2024openvla}, we expect them to inherit these structural properties.
We further conjecture that this disentanglement extends beyond task capabilities to domain knowledge. 
Specifically, we hypothesize that within a one-shot target-domain update-vector $\mathrm\Delta_{m,\text{tgt}}$, the directions capturing task-specific behavior and those capturing domain-specific adaptation can be independently identified and additively composed without mutual interference.

\vspace{-0.5em}
\subsubsection{Validation of additive composition via prototypes.}
We empirically validate the additive composition hypothesis.
Suppose we obtain a set of one-shot update-vectors $\mathrm\Delta_{m,\text{tgt}}=\theta_{m,\text{tgt}}-\theta_0$ from a set of tasks $\bm{\mathcal{T}}$ and a set of domains $\bm{\mathcal{E}}$.
From this set of update-vectors, we define three prototypes: (i) task prototype $\bar{\mathrm\Delta}_{m} := \frac{1}{|\bm{\mathcal{E}}|}\sum_{\text{tgt}} \mathrm\Delta_{m,\text{tgt}}$ averaged across domains, (ii) {domain prototype} $\bar{\mathrm\Delta}_{\text{tgt}} := \frac{1}{|\bm{\mathcal{T}}|}\sum_{m} \mathrm\Delta_{m,\text{tgt}}$ averaged across tasks, and (iii) {global prototype} $\bar{\mathrm\Delta} := \frac{1}{|\bm{\mathcal{T}}| |\bm{\mathcal{E}}|}\sum_{m,\text{tgt}} \mathrm\Delta_{m,\text{tgt}}$.
We expect that an update-vector for any specific task-domain pair $(m,\text{tgt})$ should be correctly estimated via the additive composition as $\widehat{\mathrm\Delta}_{m,\text{tgt}} := \bar{\mathrm\Delta}_{m} + \bar{\mathrm\Delta}_{\text{tgt}} - \bar{\mathrm\Delta}$.
Intuitively, $\bar{\mathrm\Delta}_{m}$ and $\bar{\mathrm\Delta}_{\text{tgt}}$ capture domain-invariant task directions and task-agnostic domain directions, respectively, and subtracting $\bar{\mathrm\Delta}$ removes common shifts inherent in fine-tuning.

We evaluate how well each prototype and composition explains $\mathrm\Delta_{m,\text{tgt}}$ using the subspace alignment metric $\gamma(\cdot,\mathrm\Delta_{m,\text{tgt}})$ in \cref{eq:sar}.
In \cref{fig:additiv_test}, the alignment scores of each prototype across 16 update-vectors exhibit low standard deviation, implying consistent task and domain directions in prototypes and thus in each update-vector. Moreover, the composed estimate $\widehat{\mathrm\Delta}_{m,\text{tgt}}$ yields the highest alignment with $\mathrm\Delta_{m,\text{tgt}}$, suggesting that domain and task adaptations map to linearly combinable directions in weight space, validating our hypothesis. 
We further discuss why the task and domain components are linearly decomposable in the supplementary material.

\vspace{-0.5em}
\subsubsection{Alignment between update-vectors on different domains.}
To further understand the properties of the domain directions in update-vectors, we keep the adaptation task fixed and evaluate the alignment of update-vectors across diverse target domains (\eg, viewpoint shifts of varying magnitude, camera noise, and their compositions) from LIBERO-Plus~\cite{fei2025liberoplus} benchmark.
The subspace alignment results in \cref{fig:heatmap_domain} reveal that the domain-specific updates learned from fine-tuning are not arbitrary, but structured.
Similar environmental changes (\eg, viewpoint shifts) produce similar update directions, while combining two shifts (\eg, viewpoint $+$ noise) produces an update that partially reuses the directions learned for each shift alone. 
This suggests the model organizes domain knowledge in a compositional way, where each type of environment change corresponds to a distinct, reusable direction in weight space.

\vspace{-0.5em}
\section{\ours: \oursFull}
\label{sec:method}

\begin{figure}[tb]
  \centering
  \includegraphics[width=\linewidth]{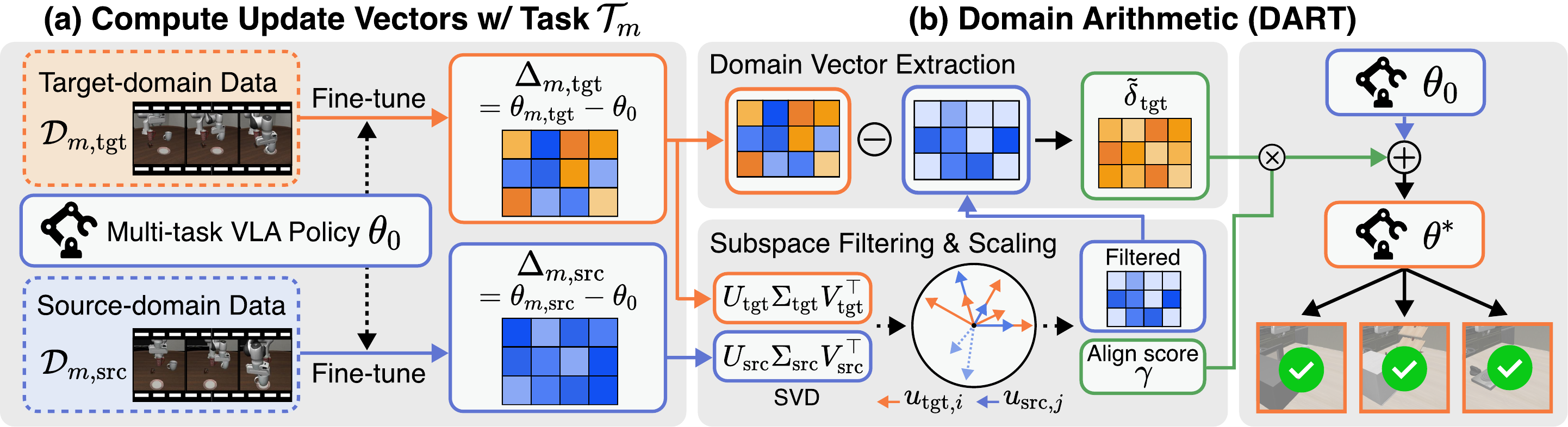}
  \caption{\textbf{Overview of the proposed VLA adaptation approach.}
  (a) We compute update-vectors $\mathrm{\Delta}_{m,src}$ and $\mathrm{\Delta}_{m,tgt}$ by fine-tuning a base policy $\theta_0$ on a single task $\mathcal{T}_m$ using source and target data. 
  (b) A domain vector $\tilde{\delta}_{tgt}$ is extracted by subtracting task directions, with subspace filtering to suppress misaligned components. 
  Adding $\tilde{\delta}_{tgt}$ back to $\theta_0$ yields a multi-task policy $\theta^*$ adapted to the target domain.}
  \label{fig:overview}
\end{figure}

In $\S$~\ref{sec:analysis}, we find that a one-shot fine-tuned model fails to adapt since its update-vector is dominated by task-relevant directions, but we also find that the update-vector can be linearly decomposable into common task and domain components.
Motivated by this, we aim to decompose and utilize the domain component for model adaptation. 
Therefore, we propose \oursFull (\textbf{\ours}), an \emph{analogy}-based method inspired by weight arithmetic~\cite{ilharco2023TA,  zhao2025adamergex}.
Instead of computing the domain prototype as in $\S$~\ref{sec:hypothesis}, which requires multiple target-domain tasks, we extract the domain direction from a single target update-vector by removing task directions using a source-domain demonstration.
To further enhance this extraction, we introduce subspace filtering and scaling that remove domain-irrelevant noise from update-vectors.
\cref{fig:overview} provides an overview of our approach.

\vspace{-0.5em}
\subsection{Domain Vector Extraction}
\label{sec:naive}

Building on the additive properties in $\S$~\ref{sec:hypothesis}, we extract the domain-specific directions by subtracting the task-specific directions from the target update-vector $\mathrm\Delta_{m,\text{tgt}}$.
To estimate these task directions without additional target-domain data, we leverage a \emph{source-domain} demonstration $\mathcal{D}_{m,\text{src}}$ of the \emph{same} task $\mathcal{T}_m$, which is typically available from the data used to train the base model $\theta_0$.
In practice, one can select $\mathcal{T}_m$ from the source dataset and then collect the corresponding target-domain expert demonstration to guarantee the same task.
Please refers to the supplementary material for detailed discussion of source-domain data.

Let $\theta^{(l)}_{m, \text{src}}$ denote the parameters for layer $l$ fine-tuned on $\mathcal{D}_{m,\text{src}}$.
Since both source update-vector $\mathrm\Delta^{(l)}_{m,\text{src}} = \theta^{(l)}_{m,\text{src}} - \theta^{(l)}_{0}$ and target update-vector $\mathrm\Delta^{(l)}_{m,\text{tgt}}$ in \cref{eq:update_vector} are learned from the same adaptation task, they primarily share common task components. 
Thus, we define the \textbf{domain vector} $\delta^{(l)}_\text{tgt}$ for layer $l$ as:

\vspace{-0.5em}
\begin{equation}\label{eq:get_domain_vector}
    \delta_\text{tgt}^{(l)} = \mathrm\Delta^{(l)}_{m,\text{tgt}} - \mathrm\Delta^{(l)}_{m,\text{src}}.
\end{equation}
This subtraction neutralizes the task-specific directions, leaving only domain-specific directions that encode the environmental shift to the target domain. 

\vspace{-0.5em}
\subsection{Subspace Alignment for Enhanced Domain Vector}
\label{sec:sar}

The domain vector extraction through direct subtraction between target and source update-vectors may successfully isolate target-domain knowledge.
However, fine-tuning often includes task-irrelevant noise~\cite{yang2025resm, yu2024dare, yadav2023ties}, and even minor source-domain artifacts can be encoded in $\mathrm\Delta_{m,\text{src}}$.
Because weight updates reside in low-rank subspaces that can be rotated or misaligned by fine-tuning artifacts~\cite{marczak2025isoc,seo2025not,zhao2024galore}, naive subtraction may inadvertently inject source-domain noise into the target domain vector or fail to remove task semantics.
To remove irrelevant noise in domain vectors, we leverage the spectral properties in one-shot update-vectors.

\vspace{-0.5em}
\subsubsection{Subspace filtering.}
Our intuition is that the shared task semantics lie in the mutually shared subspace between $\mathrm\Delta_{m,\text{src}}$ and $\mathrm\Delta_{m,\text{tgt}}$. 
By filtering the basis vectors of $\mathrm{\Delta}_{m,\text{src}}$ that weakly align with $\mathrm{\Delta}_{m,\text{tgt}}$, we can prevent source-specific noise from corrupting the domain vector.
We only filter the source update-vector as unique bases of target update-vector likely encode the domain directions we seek to isolate.
Specifically, we decompose each update-vector for layer $l$ via SVD: let $\mathrm\Delta^{(l)}_{m,\text{tgt}} = U^{(l)}_{\text{tgt}} \mathrm{\Sigma}^{(l)}_{\text{tgt}} V_{\text{tgt}}^{(l)\top}$ and $\mathrm\Delta^{(l)}_{m,\text{src}} = U^{(l)}_{\text{src}} \mathrm{\Sigma}^{(l)}_{\text{src}} V_{\text{src}}^{(l)\top}$, omitting $m$ for brevity.
We first identify which source basis vectors are geometrically aligned with the target subspace.
Following subspace alignment in model merging~\cite{marczak2025isoc, qiu2025superpose, li2026svc}, we focus on aligning the column space of the left singular vectors $U$, as it captures how the update perturbs a layer's \emph{output-feature directions} across all input directions, highly related to output changes.
We form the interaction matrix $C^{(l)} \;:=\; {U^{(l)\top}_{\text{tgt}}} U^{(l)}_{\text{src}}$ and define the overlap energy $e_j^{(l)}$ of each source basis vector $\mathbf{u}_{\text{src}, j}$ as
\begin{equation}
e^{(l)}_j \;:=\; \big\lVert C^{(l)}_{:,j} \big\rVert_2^2
\;=\
\big\lVert {U^{(l)\top}_{\text{tgt}}}^{}\mathbf{u}^{(l)}_{\text{src},j}\big\rVert_2^2,
\label{eq:overlap_energy}
\end{equation}
where $j \in \{1,\dots, R\}$  indexes the column vectors of $U_\text{src}$.
A high energy $e^{(l)}_j$ indicates that the $j$-th source basis vector lies largely within the target subspace, signifying a shared task feature. 
To retain only these shared features in $\mathrm{\Delta}^{(l)}_{m,\text{src}}$, we determine a dynamic cutoff based on the subspace alignment score $\gamma^{(l)}(\mathrm{\Delta}_{m,\text{src}},\mathrm{\Delta}_{m,\text{tgt}})$ in \cref{eq:sar}.
Since this score quantifies the fractional overlap between the two subspaces, it serves as a natural criterion for determining how many basis vectors to retain.
Let $e^{(l)}_{(1)}\ge \cdots \ge e^{(l)}_{(R)}$ denote energies sorted in descending order. We select basis vectors that are over the energy threshold by
\begin{equation}
r_l \;:=\; \min\Big\{r:\sum_{i=1}^{r} e^{(l)}_{(i)} \ge \gamma^{(l)} \sum_{j=1}^{R} e^{(l)}_{j}\Big\},
\qquad
\mathcal{J}_l \;:=\; \{\,j:\; e^{(l)}_j \ge e^{(l)}_{(r_l)} \,\},
\label{eq:greedy_select}
\end{equation}
leading to the \emph{aligned} source basis matrix $\tilde{U}^{(l)}_{\text{src}} \;:=\; U^{(l)}_{\text{src}}[:,\mathcal{J}_l]$.
We then obtain the filtered source update-vector $\tilde{\mathrm{\Delta}}^{(l)}_{m,\text{src}} = \tilde{U}^{(l)}_{\text{src}} \tilde{U}_{\text{src}}^{(l)\top} \mathrm{\Delta}^{(l)}_{m,\text{src}}$.

This subspace filtering ensures that we only subtract components that are aligned with each other. 
Unlike subspace aligning methods in model merging~\cite{marczak2025isoc, gargiulo2025tsv, yang2025resm, li2026svc}, which remove insignificant subspaces in all update-vectors and maximize each vector's unique components to maintain capabilities from each update vector, we find the common singular components between update-vectors to correctly remove common knowledge.

\vspace{-0.5em}
\subsubsection{Subspace scaling.}

Although subspace filtering can remove some misaligned singular vectors, if the two update-vectors are fundamentally misaligned, \ie, $\gamma^{(l)} \to 0$, filtering alone cannot fully ensure the correct domain vector.
Thus, we scale the domain vector by the alignment score $\gamma^{(l)}$ to down-weight if the domain vector is noise-dominant or irrelevant due to misaligned updates. 
Specifically, we obtain refined domain vector $\tilde{\delta}^{(l)}_{\text{tgt}}$ by scaling the domain vector using $\gamma^{(l)}$ as
\begin{equation}\label{eq:refined_domainvector}
\tilde{\delta}^{(l)}_{\text{tgt}} =\gamma^{(l)} \cdot \left( \mathrm{\Delta}^{(l)}_{m,\text{tgt}} -  \tilde{\mathrm{\Delta}}^{(l)}_{m,\text{src}}\right).
\end{equation}

Finally, we adapt the base policy $\theta_0$ to the target domain by injecting the domain vector into $\theta_{0}$:
\begin{equation}\label{eq:get_merged_params}
\theta^* = \theta_{0} + \alpha \cdot \tilde\delta_\text{tgt},
\end{equation}
where $\tilde\delta_\text{tgt} = \{\tilde\delta^{(l)}_\text{tgt} \}_{l=1}^{L}$ and $\alpha$ is a scalar coefficient controlling the adaptation strength.
This approach efficiently transfers the base policy to the target domain $\mathcal{E}_{\text{tgt}}$ while preserving the multi-task capabilities inherent in $\theta_{0}$.

\section{Experiments}\label{sec:experiments}

\vspace{-0.5em}
To evaluate our method, we conduct (i) simulation experiments under diverse visual shifts and cross-embodiment transfer ($\S$~\ref{subsec:sim_results}), (ii) real-world experiments under viewpoint shifts ($\S$~\ref{subsec:realworld_results}), and (iii) additional analyses ($\S$~\ref{subsec:analysis}).

\vspace{-0.5em}
\subsection{Setups}\label{subsec:setups}

\begin{figure}[tb]
    \centering
    \includegraphics[width=\linewidth]{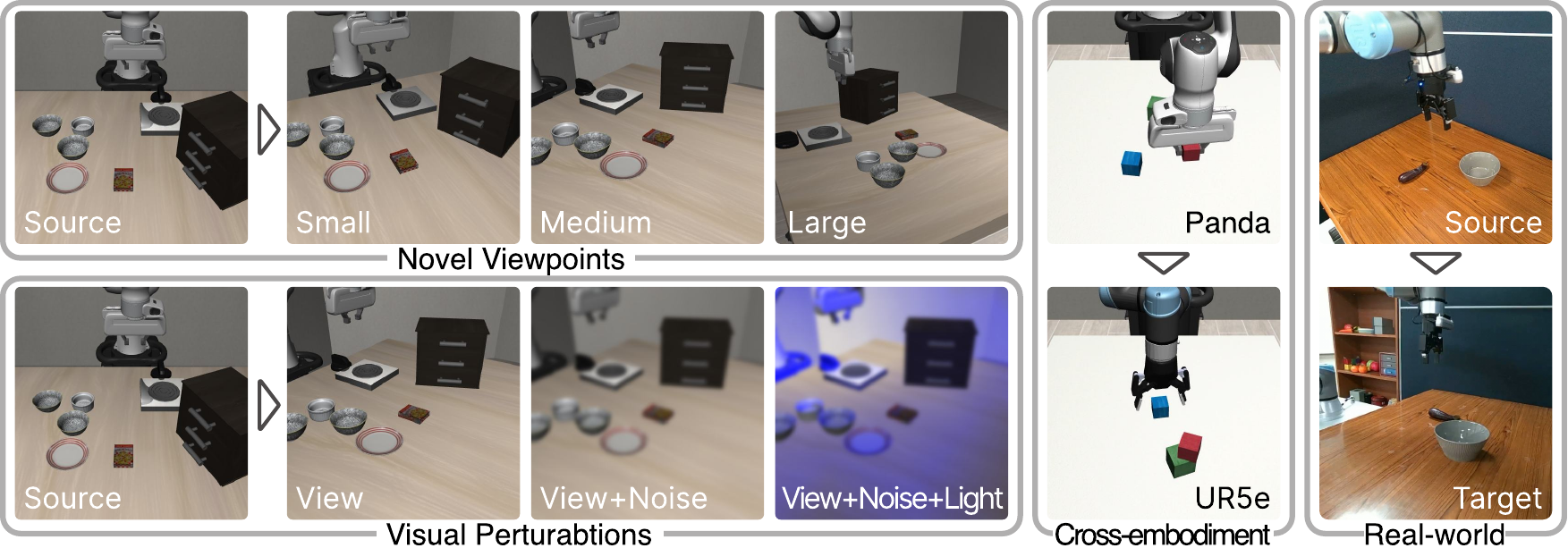}
    \caption{\textbf{Overview of experimental setups.} We experiment on four setups: simulation setups with novel viewpoints (top-left) and combined visual perturbations (bottom-left) on LIBERO~\cite{liu2023libero}, a cross-embodiment transfer setup on MimicGen~\cite{mandlekar2023mimicgen} (middle), and a real-world setup on two third-person camera viewpoints (right).}
    \label{fig:simulation_setup}
    \vspace{-0.5em}
\end{figure}

\subsubsection{Models.}
We primarily evaluate our approach on \piOfive~\cite{zhou2025pi05}, a flow-matching-based VLA model~\cite{lipman2022flowmatching}. 
To assess architectural generality, we additionally evaluate on \pifast~\cite{pertsch2025fast}, which uses autoregressive action-token generation. 

\vspace{-0.5em}
\subsubsection{Baselines.}
In both simulation and real-world experiments, we compare \ours with architecture-agnostic VLA adaptation baselines: 
(i) \textbf{Zero-shot} (no adaptation), 
(ii) \textbf{One-shot FT} (full fine-tuning on the one-shot dataset), 
(iii) \textbf{FLA}~\cite{li2025fla} (vision encoder adaptation using LoRA~\cite{hu2022lora}), and 
(iv) \textbf{RETAIN}~\cite{yadav2025retain} (model merging between source model and One-shot FT with module-wise scaling).

\vspace{-0.5em}
\subsubsection{Simulation setup.}
For visual shifts, we evaluate on LIBERO~\cite{liu2023libero}, a robot manipulation benchmark with four task suites of total 40 tasks.
Following prior work~\cite{li2025fla, wilcox2025adapt3r, fei2025liberoplus}, we apply third-person viewpoint shifts and visual perturbations to LIBERO to mimic real-world environmental shifts (\cref{fig:simulation_setup} left). 
We consider three levels of viewpoint shift relative to the source camera pose: \texttt{Small}, \texttt{Medium}, and \texttt{Large}, and two visual perturbations applied on top of the viewpoint shifts: \texttt{Noise} (noise injection) and \texttt{Light} (illumination change).
We adapt the base model trained on the original LIBERO training dataset~\cite{zhou2025pi05, pertsch2025fast} to each target domain using a scene-wise one-shot dataset collected in the target domain, containing a single demonstration from one task in each of the five LIBERO scenes.
We repeat one-shot adaptation three times with different randomly selected adaptation tasks and report the average \emph{Success Rate} (\%), with 50 rollouts per task.

For cross-embodiment transfer, we evaluate on MimicGen~\cite{mandlekar2023mimicgen} with \texttt{Stack} and \texttt{Stack Three} tasks, transferring from Panda to UR5e (\cref{fig:simulation_setup} middle).
Based on MimicGen-pretrained \piOfive, the base policy $\theta_0$ is trained on two tasks with Panda.
We then derive the source- and target-domain vector using one \texttt{Stack} demonstration from each robot.
We report \emph{Progress Rate} (Prog., \%) and \emph{Success Rate} (Succ., \%) as metrics, averaged over 5 seeds with 50 rollouts per seed.

\vspace{-0.5em}
\subsubsection{Real-world setup.}
We evaluate on five real-world tasks using a 6-DoF UR10e robot arm with a Robotiq 2F-85 gripper (\cref{fig:simulation_setup} right): three pick-and-place tasks (\texttt{Eggplant}, \texttt{Lemon}, \texttt{Carrot}) and two fine-grained manipulation tasks (\texttt{Stack Cube}, \texttt{Press Stapler}).
We use 120 demonstrations (24 per task) collected under the \texttt{Source} viewpoint to train \piOfive, and collect one additional \texttt{Stack Cube} demonstration under the \texttt{Target} viewpoint for adaptation.
We report \emph{Success Rate} (\%), averaged over 12 rollouts per task with distinct object placements.

\begin{table}[tb!]
\centering
\caption{\textbf{Performance on LIBERO across novel viewpoints using \piOfive.} 
We report average success rates of total 40 tasks for three trials of one-shot adaptation with different adaptation tasks, with the best in \textbf{bold}.}
\vspace{-0.5em}
\label{tab:main_results}

\renewcommand{\arraystretch}{1.0}
\setlength{\tabcolsep}{5pt}

\begin{tabularx}{\linewidth}{l *{4}{>{\centering\arraybackslash}X}}
\toprule
 & \multicolumn{4}{c}{Novel Viewpoints (Success Rate, \%)} \\
\cmidrule(lr){2-5}
Methods (\piOfive) 
& \texttt{Small} 
& \texttt{Medium} 
& \texttt{Large} 
& Average \\
\midrule
Zero-shot & 88.3 & 63.9 & 11.3 & 54.5 \\
One-shot FT & 43.4 & 33.3 & 17.8 & 31.5 \\
RETAIN \textcolor{blue}{(ICLR 2026)} & 87.4 & 72.4 & 48.9 & 69.6 \\
FLA \textcolor{blue}{(CVPR 2026)} & \textbf{92.2} & 76.4 & 54.3 & 74.3 \\
\rowcolor{gray!15}
\ours (\textbf{Ours}) & 92.0 & \textbf{80.8} & \textbf{64.4} & \textbf{79.1} \\
\bottomrule
\end{tabularx}
\end{table}

\begin{table}[!t]
\centering
\caption{\textbf{Performance on LIBERO under combined visual shifts using \piOfive.}
We evaluate under the \texttt{Medium} viewpoint shift (\texttt{View}) and two combined settings: \texttt{View+Noise} (camera noise) and \texttt{View+Noise+Light} (camera noise with illumination).}
\vspace{-0.5em}
\label{tab:visual_perturbations}

\renewcommand{\arraystretch}{1.0}
\setlength{\tabcolsep}{4pt}

\begin{tabularx}{\linewidth}{
l
>{\centering\arraybackslash}X
>{\centering\arraybackslash}X
>{\centering\arraybackslash}p{0.25\linewidth}
>{\centering\arraybackslash}p{0.12\linewidth}
}
\toprule
 & \multicolumn{4}{c}{Visual Perturbations (Success Rate, \%)} \\
\cmidrule(lr){2-5}
Methods (\piOfive)
& \texttt{View}
& \texttt{View+Noise}
& \texttt{View+Noise+Light}
& Average \\
\midrule
Zero-shot & 63.9 & 60.3 & 57.2 & 60.5 \\
One-shot FT & 33.3 & 27.7 & 28.5 & 29.8 \\
RETAIN \textcolor{blue}{(ICLR 2026)} & 72.4 & 65.2 & 68.5 & 68.7 \\
FLA \textcolor{blue}{(CVPR 2026)} & 76.4 & 67.8 & 70.2 & 71.5 \\
\rowcolor{gray!15}
\ours (\textbf{Ours}) & \textbf{80.8} & \textbf{69.2} & \textbf{75.0} & \textbf{75.0} \\
\bottomrule
\end{tabularx}
\vspace{-1.0em}
\end{table}

\vspace{-0.5em}
\subsubsection{Implementation details.}

For \ours, we fine-tune source and target one-shot models ($\theta_{m,\text{src}}, \theta_{m,\text{tgt}}$) for $1{,}000$ steps from the base model $\theta_0$.
We set the scaling coefficient $\alpha$ to $0.8$ for \ours via a small search (10 rollouts per task) on the \texttt{Medium} viewpoint in a single task suite of LIBERO, and use the same value for all other task suites, viewpoints, architectures, and in the real-world experiments.
We use the same procedure to find hyperparameters and evaluate each baseline on the same one-shot dataset as our method.
Additional details of experimental setup are provided in the supplementary material. 

\vspace{-0.5em}
\subsection{Simulation Results}
\label{subsec:sim_results}

\subsubsection{Novel visual domains.}
As shown in~\cref{tab:main_results}, \ours outperforms all baselines under diverse novel viewpoints.
Notably, applying the domain vector $\tilde\delta_\text{tgt}$ to the base policy ($\theta_{0}$) yields a substantial gain of 24.6 percentage points (pp).
These results support our hypothesis that domain vectors provide domain-specific knowledge while preserving multi-task capabilities.
Our method outperforms FLA~\cite{li2025fla}, which fine-tunes only the vision encoder, highlighting the importance of adapting the entire model in a data-limited setting.
Compared with RETAIN~\cite{yadav2025retain}, our analogy-based approach performs better, suggesting that explicitly isolating domain-shift directions helps under data scarcity.

\cref{tab:visual_perturbations} summarizes performance under combined visual perturbations.
\ours maintains a clear advantage over baselines across all settings.
This trend indicates that the advantage of our method persists even under combined environmental shifts, rather than being limited to a single perturbation type.

\begin{table}[!t]
\centering
\caption{\textbf{Performance on LIBERO across novel viewpoints using \pifast.}
We report average success rates of total 40 tasks for three trials of one-shot adaptation with different adaptation tasks, with the best in \textbf{bold}.}
\vspace{-0.5em}
\label{tab:pifast_results}

\renewcommand{\arraystretch}{1.0}
\setlength{\tabcolsep}{6pt}

\begin{tabularx}{\linewidth}{l *{4}{>{\centering\arraybackslash}X}}
\toprule
 & \multicolumn{4}{c}{Novel Viewpoints (Success Rate, \%)} \\
\cmidrule(lr){2-5}
Methods (\pifast) & \texttt{Small} & \texttt{Medium} & \texttt{Large} & Average \\
\midrule
Zero-shot & 84.6 & 73.6 & 62.0 & 73.4 \\
One-shot FT & 71.1 & 63.0 & 52.2 & 62.1 \\
RETAIN \textcolor{blue}{(ICLR 2026)} & 88.3 & 78.4 & 62.7 & 76.5 \\
FLA \textcolor{blue}{(CVPR 2026)} & 86.5 & 78.4 & 64.9 & 76.6 \\
\rowcolor{gray!15}
\ours (\textbf{Ours}) & \textbf{91.2} & \textbf{80.8} & \textbf{66.2} & \textbf{79.4} \\
\bottomrule
\end{tabularx}
\end{table}

\begin{table}[!t]
\centering
\caption{\textbf{Performance on MimicGen under cross-embodiment transfer using $\pi_{0.5}$.}
We adapt a source policy trained on the Panda robot to the UR5e robot, and report progress rate and success rate.}
\vspace{-0.5em}
\label{tab:cross_embodiment_results}

\renewcommand{\arraystretch}{1.0}
\setlength{\tabcolsep}{3.5pt}

\begin{tabularx}{\linewidth}{l *{6}{>{\centering\arraybackslash}X}}
\toprule
 & \multicolumn{2}{c}{\texttt{Stack}}
 & \multicolumn{2}{c}{\texttt{Stack Three}}
 & \multicolumn{2}{c}{Average} \\
\cmidrule(lr){2-3}
\cmidrule(lr){4-5}
\cmidrule(lr){6-7}
Methods (\piOfive)
& Prog. (\%) & Succ. (\%)
& Prog. (\%) & Succ. (\%)
& Prog. (\%) & Succ. (\%) \\
\midrule
Zero-shot & 89.4 & 86.8 & 70.1 & 37.2 & 79.8 & 62.0 \\
One-shot FT & 87.8 & 84.8 & 60.9 & 28.0 & 74.4 & 56.4 \\
\rowcolor{gray!15}
\ours (\textbf{Ours}) & \textbf{94.8} & \textbf{93.4} & \textbf{73.8} & \textbf{45.4} & \textbf{84.3} & \textbf{69.4} \\
\bottomrule
\end{tabularx}
\vspace{-0.75em}
\end{table}

\vspace{-0.5em}
\subsubsection{Applicability to an alternative VLA architecture.}
To test generalizability beyond \piOfive's flow-matching formulation, we apply our approach to \pifast, an autoregressive VLA model. 
As shown in~\cref{tab:pifast_results}, our method also consistently outperforms all baselines on this architecture.
This result supports that the additive structure of update-vectors is applicable to diverse VLA architectures.

\vspace{-0.5em}
\subsubsection{Cross-embodiment transfer.}
Beyond visual domain shifts, upgrading hardware or transferring policies across different robotic platforms remains a major bottleneck in real-world deployment.
To test if our approach can bridge this physical domain gap, we examine whether the analogy principle in~\cref{eq:get_domain_vector} extends to cross-embodiment transfer.
\cref{tab:cross_embodiment_results} shows that \ours is also applicable to the cross-embodiment setting, which differs substantially from visual domain adaptation.
This highlights that our approach can be applied across diverse visual and physical environmental shifts without any algorithmic modification.

\vspace{-0.5em}
\subsection{Real-world Results}\label{subsec:realworld_results}

\begin{table}[!t]
\centering
\caption{\textbf{Performance on real-world UR10e robot using $\pi_{0.5}$.}
We use a single \texttt{Stack Cube} demonstration to adapt models to the target domain (novel viewpoint).}
\vspace{-0.5em}
\label{tab:realworld}

\setlength{\tabcolsep}{4pt}
\renewcommand{\arraystretch}{1.0}

\resizebox{\columnwidth}{!}{%
\begin{tabular}{lcccccc}
\toprule
 & \multicolumn{6}{c}{Novel Viewpoint (Success Rate, \%)} \\
\cmidrule(lr){2-7}
 & \multicolumn{3}{c}{Pick-and-Place} & \multicolumn{2}{c}{Fine-grained Manipulation} & \\
\cmidrule(lr){2-4}\cmidrule(lr){5-6}
Methods ($\pi_{0.5}$)
& \texttt{Eggplant} & \texttt{Lemon} & \texttt{Carrot}
& \texttt{Stack Cube} & \texttt{Press Stapler} & Average \\
\midrule
Zero-shot & 50.0 & 33.3 & 41.7 & 16.7 & 75.0 & 43.3 \\
One-shot FT & 58.3 & 58.3 & 41.7 & 33.3 & 66.7 & 51.7 \\
RETAIN \textcolor{blue}{(ICLR 2026)} & 58.3 & 41.7 & 41.7 & 16.7 & 83.3 & 48.3 \\
FLA \textcolor{blue}{(CVPR 2026)} & 58.3 & 50.0 & 50.0 & 16.7 & \textbf{100.0} & 55.0 \\
\rowcolor{gray!15}
\ours (\textbf{Ours})
& \textbf{91.7} & \textbf{91.7} & \textbf{83.3}
& \textbf{41.7} & \textbf{100.0} & \textbf{81.7} \\
\bottomrule
\end{tabular}%
}
\vspace{-1.0em}
\end{table}

We evaluate \ours in a real-world setup to assess whether it remains effective under real-world variability.
\Cref{tab:realworld} summarizes results under third-person camera viewpoint shifts on a UR10e robot.
Despite adapting from only a single \texttt{Stack Cube} demonstration, our method achieves high success rates across all five tasks, indicating that their domain-level transfer is effective even in diverse real-world conditions. 
In contrast, baselines perform substantially worse under the same one-shot budget, indicating limited transfer beyond the demonstration.
We provide experiment videos in the supplementary material.

\vspace{-0.5em}
\subsection{Detailed Analysis}
\label{subsec:analysis}

\subsubsection{Ablation study.}
\cref{tab:ablation} isolates the effect of each component in \ours. 
\ours without any subspace alignment component yields substantial improvement over One-shot FT (in \cref{tab:main_results}), suggesting that the analogy-based weight arithmetic between source- and target-domain update vectors effectively isolates domain-specific knowledge. 
This validates the predominant, domain-agnostic task-specific directions among update vectors evidenced in §\ref{sec:similarity}, and confirms their linearly decomposable property consistent with §\ref{sec:hypothesis}. 
Building on this, subspace filtering leads to marked improvement, highlighting the importance of suppressing noisy source-domain artifacts in the source-domain update vector for more precise domain vector extraction. 
Furthermore, subspace scaling yields additional gains, suggesting the presence of highly misaligned, low-quality domain vectors whose contribution is better modulated through alignment-aware reweighting.

\vspace{-1.0em}
\subsubsection{Merging multiple domain vectors.}

We study whether domain vectors for different target domains from \ours can be consolidated into a single transferable vector, motivated by the additivity of domain directions in weight space ($\S$~\ref{sec:hypothesis}). 
We merge three novel viewpoint-shift domain vectors $\tilde\delta_{\text{tgt}}$ from LIBERO into a combined vector $\delta^{*}$ using model-merging methods~\cite{ilharco2023TA, yadav2023ties, gargiulo2025tsv, marczak2025isoc}, and adapt the base model as $\theta^* = \theta_{0} + \alpha \cdot \delta^{*}$. 
As shown in \cref{tab:merge_domain_vectors}, the merged vector successfully adapts the base model to all three domains, emphasizing the composable nature of domain vectors. 
This indicates the practicality of \ours that only a single consolidated vector can be maintained across multiple target domains, reducing the memory overhead of storing each domain vector.

\begin{figure}[t!]
  \centering
  \begin{minipage}[t]{0.48\linewidth}
    \centering
    \vspace{0pt}
    \captionof{table}{\textbf{Ablation study of each component in \ours.}
    We report average success rates (\%) across \texttt{Small}, \texttt{Medium}, and \texttt{Large} viewpoints in LIBERO.
    \texttt{Sub. Filter.} is subspace filtering, and \texttt{Sub. Scale.} is subspace scaling in $\S$~\ref{sec:sar}.}
    \label{tab:ablation}
    \vspace{0.5em}
    \setlength{\tabcolsep}{4.0pt}
    \begin{tabular}{ccc}
      \toprule
      \multicolumn{2}{c}{Components} & \\
      \cmidrule(lr){1-2}
      \texttt{Sub. Filter.} &
      \texttt{Sub. Scale.} &
      Average \\
      \midrule
      \XSolidBrush & \XSolidBrush & 78.1 \\
      \Checkmark   & \XSolidBrush & 78.8 \\
      \XSolidBrush & \Checkmark   & 78.5 \\
      \Checkmark   & \Checkmark   & \textbf{79.1} \\
      \bottomrule
    \end{tabular}
  \end{minipage}
  \hfill
  \begin{minipage}[t]{0.48\linewidth}
  \centering
  \vspace{0pt}
  \centering
\captionof{table}{\textbf{Merging domain vectors $\tilde\delta_\text{tgt}$ across novel viewpoints.}
\ours reports average success rates (\%) of a separately adapted model in three novel viewpoints in LIBERO. Each single merged model is evaluated across the three domains.
}
\label{tab:merge_domain_vectors}
\vspace{0.5em}
\setlength{\tabcolsep}{8pt}
\begin{tabular}{llc}
\toprule
\multicolumn{2}{l}{Methods (\piOfive)} & Average \\
\midrule
\multicolumn{2}{l}{\ours} & \textbf{79.1} \\
\midrule
\multirow{4}{*}{\makecell[l]{\ours\\$+$ Merging}}
  & TA~\cite{ilharco2023TA}  & 74.5 \\
  & TIES~\cite{yadav2023ties} & 70.9 \\
  & TSV~\cite{gargiulo2025tsv} & 75.7 \\
  & Iso-C~\cite{marczak2025isoc} & 74.8 \\
\bottomrule
\end{tabular}
      
  \end{minipage}
\end{figure}

\begin{figure}[t!]
  \vspace{-0.5em}
  \centering
  \begin{subfigure}{0.49\linewidth}
    \includegraphics[width=0.99\linewidth]{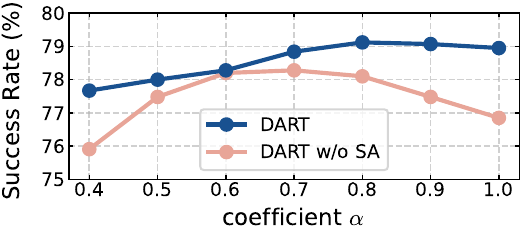}
    \caption{\textbf{Impact of scaling coefficient $\alpha$.}
    }
    \label{fig:effect_hyperparam}
  \end{subfigure}
  \hfill
  \begin{subfigure}{0.49\linewidth}
    \includegraphics[width=0.99\linewidth]{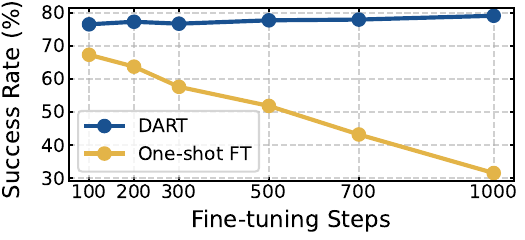}
    \caption{\textbf{Impact of fine-tuning steps.}
    }
    \label{fig:effect_iteration}
  \end{subfigure}
  \caption{\textbf{Performance under hyperparameter choices on LIBERO across novel viewpoints.}
  We average success rates (\%) across \texttt{Small}, \texttt{Medium}, \texttt{Large} camera views.
  }
  \vspace{-0.5em}
  \label{fig:hyperparam_total}
\end{figure}

\vspace{-1.0em}
\subsubsection{Scaling coefficient $\alpha$.}
\Cref{fig:effect_hyperparam} shows how \ours and its simplified version, \ours without Subspace Alignment (\ours w/o SA) that does not apply subspace filtering or scaling, vary in performance with the scaling coefficient $\alpha$ in~\cref{eq:get_merged_params}.
Small $\alpha$ is insufficient to address the environmental shift, whereas large $\alpha$ can interfere with the base policy’s multi-task capabilities. 
Nevertheless, \ours maintains strong performance across a wide range of $\alpha$. 
In particular, \ours is more stable than \ours w/o SA, suggesting that subspace filtering and scaling suppress noisy components that would otherwise be amplified by $\alpha$.

\vspace{-1.0em}
\subsubsection{Fine-tuning steps.}
\cref{fig:effect_iteration} shows the effect of the number of fine-tuning steps for the update-vectors in~\cref{eq:get_domain_vector}.
While One-shot FT degrades over time due to catastrophic forgetting~\cite{yadav2025retain}, \ours, which extracts the domain vector from One-shot FT weights, shows small but consistent performance gains, likely as task-specific directions become more pronounced with training, leading to better domain vector extraction.
Notably, \ours maintains strong performance even at small fine-tuning steps, enabling time-efficient adaptation.
 
We further show that \ours avoids source-domain forgetting, compare \ours with existing model-merging and test-time-adaptation methods, and analyze the choice of layers to which we add the domain vector and the choice of tasks used to extract it.
Please see the supplementary material for details.

\vspace{-0.5em}
\section{Limitation}
While DART consistently outperforms baselines, performance degrades under severe shifts (e.g., \texttt{Large} viewpoint), a challenge shared by all one-shot methods. We leave this to future work, \eg, via more reliable domain vector extraction or stronger base-model training/fine-tuning schemes.
Additionally, the scalar coefficient $\alpha$ requires a small hyperparameter search, though our analysis shows DART remains stable across a wide range of values and generalizes well across environments. Hyperparameter-free per-layer adaptive scaling for practical real-world application is left for future work.

\vspace{-0.5em}
\section{Conclusion}

We propose a method to adapt VLA models for environmental shifts with only a single demonstration collection. 
Motivated by our observation that one-shot fine-tuned parameters admit an approximately additive decomposition into task- and domain-specific directions,
we introduce \ours, an analogy-based approach that adds filtered domain-specific directions isolated by weight arithmetic.
Extensive evaluation in simulation and on real-world setups shows consistent improvement and applicability of \ours across diverse visual and embodiment shifts.


\section*{Acknowledgements}
We thank Seongwon Cho, Youhan Lee, and Jimin Nam for their helpful comments.
This work was partly supported by the InnoCORE program (26-InnoCORE-01), the IITP grants (RS-2022-II220077, RS-2022-II220113, RS-2022-II220959, RS-2022-II220871, RS-2026-25507282, RS-2026-25518317, RS-2021-II211343 (SNU AI), RS-2025-25442338 (AI Star Fellowship-SNU)), 02-26-01-0285 (Advanced GPU Utilization Support Program by NIPA) funded by the Korea government (MSIT), grants (RS-2025-25462891 (US-KOR BARI), RS-2025-25453780) funded by MOTIR, a grant (RS-2025-25460896) funded by MOTIR and KIAT, a grant of Korean ARPA-H Project through the Korea Health Industry Development Institute (KHIDI), funded by the Ministry of Health \& Welfare, Republic of Korea (RS-2025-25424639), and the BK21 FOUR program, SNU in 2025.

%
%
\bibliographystyle{splncs04}
\bibliography{main}


\clearpage

\appendix
\noindent{\large\bfseries Supplementary Material}
\vspace{0.5em}

This supplementary material provides additional technical details, experimental protocols, and extended analyses for \ours. Specifically, we include:

\paragraph{\textbf{Method Details.}}
\begin{itemize}
\vspace{-0.5em}
    \item \textbf{\Cref{appx:algorithm}:} Overall algorithm of \ours.
    \item \textbf{\Cref{appx:justification_source_demo}:} Justification for using source-domain demonstrations.
    \item \textbf{\Cref{appx:efficient_svd}:} Acceleration of \ours with randomized SVD.
    \item \textbf{\Cref{appx:why_decompose}:} Motivation for task--domain decomposition in update-vectors.
\end{itemize}

\paragraph{\textbf{Baseline Details.}}
\begin{itemize}
\vspace{-0.5em}
    \item \textbf{\Cref{appx:baseline_retain}:} Implementation details of RETAIN.
    \item \textbf{\Cref{appx:baseline_fla}:} Implementation details of FLA.
\end{itemize}

\paragraph{\textbf{Experimental Setup.}}
\begin{itemize}
\vspace{-0.5em}
    \item \textbf{\Cref{appx:hyperparam}:} VLA model and training hyperparameters.
    \item \textbf{\Cref{appx:libero_setup}:} LIBERO setup for visual domain shifts.
    \item \textbf{\Cref{appx:mimicgen_setup}:} MimicGen setup for cross-embodiment transfer.
    \item \textbf{\Cref{appx:realworld_setup}:} Real-world robot setup.
\end{itemize}

\paragraph{\textbf{Additional Results and Analysis.}}
\begin{itemize}
\vspace{-0.5em}
    \item \textbf{\Cref{appx:libero_suitewise}:} Suite-wise LIBERO results.
    \item \textbf{\Cref{appx:upperbound}.:} Full-data fine-tuning upper-bound results.
    \item \textbf{\Cref{appx:model_merging_methods}:} Comparison with model merging methods.
    \item \textbf{\Cref{appx:test_time_adaptation}:} Comparison with test-time adaptation.
    \item \textbf{\Cref{appx:detailed_analysis}:} Further analysis of \ours.
\end{itemize}

  
\vspace{-0.5em}
\section{Details of \ours}
\label{appx:detail_ours}

\vspace{-0.5em}
\subsection{Algorithm}
\label{appx:algorithm}

We summarize the procedure of our analogy-based proposed method, \oursFull (\textbf{\ours}), in \cref{alg:dart}.

\begin{algorithm}[!]
\caption{\textsc{DART}: Domain Arithmetic for One-shot VLA Adaptation}
\label{alg:dart}
\begin{algorithmic}[1]
\State \textbf{Input:} Base multi-task policy parameters $\theta_{0}$, adapt-task index $m$, train datasets
$\mathcal{D}_{m,\text{src}}$ (source-domain) and $\mathcal{D}_{m,\text{tgt}}$ (target-domain),
scaling coefficient $\alpha$
\State \textbf{Output:} Adapted parameters $\theta^{*}$

\Statex
\State \textit{// (a) Compute one-shot update-vectors}
\State $\theta_{m,\text{src}} \gets \textsc{FineTune}(\theta_{\text{src}}, \mathcal{D}_{m,\text{src}})$
\State $\theta_{m,\text{tgt}} \gets \textsc{FineTune}(\theta_{\text{src}}, \mathcal{D}_{m,\text{tgt}})$
\State $\mathrm\Delta_{m,\text{src}} \gets \theta_{m,\text{src}} - \theta_{\text{src}}$
\State $\mathrm\Delta_{m,\text{tgt}} \gets \theta_{m,\text{tgt}} - \theta_{\text{src}}$

\Statex
\State \textit{// (b) Domain vector extraction in aligned subspace}
\For{$l = 1$ \textbf{to} $L$} 
\If{layer is not 2-D, \eg, bias or norm} \Comment{for non-linear layers}
  \State $\tilde{\delta}^{(l)}_{\text{tgt}} \gets
  \mathrm\Delta^{(l)}_{m,\text{tgt}} - \mathrm\Delta^{(l)}_{m,\text{src}}$
\Else
\State \textit{// SVD}
\State $\mathrm\Delta^{(l)}_{m,\text{tgt}} = U^{(l)}_{\text{tgt}} \Sigma^{(l)}_{\text{tgt}} {V^{(l)}_{\text{tgt}}}^{\top}$
\State $\mathrm\Delta^{(l)}_{m,\text{src}} = U^{(l)}_{\text{src}} \Sigma^{(l)}_{\text{src}} {V^{(l)}_{\text{src}}}^{\top}$

\State \textit{// Subspace alignment score $\gamma^{(l)}(\mathrm\Delta_{m,\text{src}},\mathrm\Delta_{m,\text{tgt}})$ (Eq.\ 2)}
\State $\gamma^{(l)} \gets \dfrac{\left\lVert U^{(l)}_{\text{tgt}} {U^{(l)}_{\text{tgt}}}^{\top} \,\mathrm\Delta^{(l)}_{m,\text{src}} \right\rVert_F}{\left\lVert \mathrm\Delta^{(l)}_{m,\text{src}} \right\rVert_F}$

\State \textit{// Overlap energy for each source basis (Eq.\ 4)}
\State $C^{(l)} \gets {U^{(l)}_{\text{tgt}}}^{\top} U^{(l)}_{\text{src}}$
\For{$j = 1$ \textbf{to} $R$}
  \State $e^{(l)}_j \gets \left\lVert C^{(l)}_{:,j} \right\rVert_2^2$
\EndFor

\State \textit{// Greedy selection threshold using $\gamma^{(l)}$ (Eq.\ 5)}
\State Sort $\{e^{(l)}_j\}_{j=1}^{R}$ in descending order to get $e^{(l)}_{(1)} \ge \cdots \ge e^{(l)}_{(R)}$
\State $r_l \gets \min\Big\{r:\sum_{i=1}^{r} e^{(l)}_{(i)} \ge \gamma^{(l)} \sum_{j=1}^{R} e^{(l)}_{j}\Big\}$
\State $\mathcal{J}_l \gets \{\, j:\; e^{(l)}_j \ge e^{(l)}_{(r_l)} \,\}$
\State $\tilde{U}^{(l)}_{\text{src}} \gets U^{(l)}_{\text{src}}[:,\mathcal{J}_l]$

\State \textit{// Refined domain vector (Eq.\ 6)}
\State $\tilde{\delta}^{(l)}_{\text{tgt}} \gets \gamma^{(l)} \cdot
  \Big(\mathrm\Delta^{(l)}_{m,\text{tgt}} - \tilde{U}^{(l)}_{\text{src}} {\tilde{U}^{(l) \top}_{\text{src}}} \mathrm\Delta^{(l)}_{m,\text{src}}\Big)$
\EndIf
\EndFor

\Statex
\State \textit{// Adapt the multi-task policy by adding the domain vector (Eq.\ 7)}
  \State $\theta^{*} \gets \theta_{0} + \alpha \cdot \tilde{\delta}_{\text{tgt}}$

\State \Return $\theta^{*}$
\end{algorithmic}
\end{algorithm}

\vspace{-0.5em}
\subsection{Justification for Using Source-domain Demonstrations}
\label{appx:justification_source_demo}

A key assumption in \ours is access to the source-domain training dataset used to pretrain the source policy, along with at least one source-domain demonstration for the adaptation task $\mathcal{T}_m$. Below we explain why this assumption is plausible and how it can be relaxed.

\vspace{-0.5em}
\subsubsection{Availability of source-domain demonstrations.}
In many robotics settings, pretrained policies are trained on large-scale open-source robotic datasets~\cite{o2024oxe,khazatsky2024droid}, from which we can retrieve demonstrations for adaptation.
Also, since recent VLA models commonly rely on task-wise fine-tuning on teleoperated demonstrations to perform multiple tasks in a given environment~\cite{kim2024openvla,kim2025oft,zhou2025pi05}, it is natural that the demonstrations used for such fine-tuning are available as source data.
Moreover, our method requires only \emph{a small number} of source-domain demonstrations (e.g., a single trajectory) to compute $\mathrm\Delta_{m,\text{src}}$, rather than access to the full dataset.

\vspace{-0.5em}
\subsubsection{Obtaining the same adaptation task across domains.}
The assumption that we can identify the same task $\mathcal{T}_m$ in the source dataset is straightforward when the source training set contains a limited and well-defined task taxonomy (as in standard benchmarks). When the source dataset is large or weakly organized, an exact task lookup may be difficult.
However, this does not prevent our approach in practice: we can instead \emph{choose} the adaptation task from the source side first. 
Concretely, we sample a source-domain demonstration from the training set, treat its underlying task as $\mathcal{T}_m$, and then collect a target-domain expert demonstration for the \emph{same} task. This simple protocol guarantees the same adaptation task by construction, avoiding the need for explicit task indexing in the source dataset.

\vspace{-0.5em}
\subsubsection{Robustness to imperfect task matching.}
Even when exact matching is not possible (\eg, when we can only collect expert demonstration on certain tasks), our method remains usable. In \cref{tab:src_tgt_combination}, we show that using a source update-vector from a \emph{different} task ($m'\neq m$) degrades performance, but selecting a \emph{similar} task (via feature cosine similarity) consistently outperforms a random choice. This suggests that approximate task matching can still yield meaningful domain vectors, and that performance can further improve with stronger task retrieval mechanisms~\cite{xie2025iwr,dass2025datamil,kumar2025collage} and with more diverse source training sets~\cite{o2024oxe,khazatsky2024droid,walke2023bridgedata}.

\vspace{-0.5em}
\subsubsection{Takeaway.}
Overall, requiring a source-domain demonstration is a mild and practical assumption: it can be satisfied either by direct access to source training data, or by selecting the adaptation task from the source dataset and collecting the corresponding target-domain demonstration.
When only approximate matches are available, similarity-based retrieval provides a viable fallback with room for improvement.

\vspace{-0.5em}
\subsection{Accelerating \ours with Randomized SVD}
\label{appx:efficient_svd}

\begin{table}[t]
\centering
\caption{
    \textbf{Effect of randomized SVD.}
    We report success rate (\%) and runtime on novel viewpoints.
    For randomized SVD, we use a target rank of $r{=}256$.
    Runtime is averaged over three runs and measured on a machine with 1TB RAM and an Intel Xeon Platinum 8562Y+ CPU.
}
\label{tab:randomized_svd}
\begin{tabular*}{0.9\columnwidth}{@{\extracolsep{\fill}}lccccc@{}}
\toprule
 & \multicolumn{4}{c}{Novel Viewpoints (Success Rate, \%)} & \\
\cmidrule(lr){2-5}
Method (\piOfive) & \texttt{Small} & \texttt{Medium} & \texttt{Large} & Average & Runtime \\
\midrule
\ours $+$ Full SVD         & 92.0  & 80.8  & 64.4  & 79.1 & 15m 35s \\
\ours $+$ Randomized SVD   & 91.7  & 80.7  & 63.8  & 78.7 & 6m 33s  \\
\bottomrule
\end{tabular*}
\vspace{-0.5em}
\end{table}

\vspace{-0.5em}
Our method is computationally lightweight compared to training-based adaptation baselines~\cite{fei2025liberoplus,li2025fla}, as it only requires a few one-shot fine-tuning runs followed by weight-space arithmetic.
However, \ours involves computing Singular Value Decompositions (SVD) of both base and target update vectors at each layer, which can be computationally expensive for larger models.
To reduce this overhead, we replace full SVD with a truncated randomized SVD approximation~\cite{halko2011randomizedsvd} (Randomized SVD), which estimates the top-$r$ singular subspace without performing a full decomposition.
This reduces the dominant per-layer cost from $O(m n \min(m,n))$ (full SVD) to approximately $O(m n r)$ for an $m\times n$ matrix with target rank $r \ll \min(m,n)$, where $m$ and $n$ denote the output and input dimensions of the layer.
As shown in~\cref{tab:randomized_svd}, Randomized SVD achieves comparable performance to full SVD while substantially reducing computation.

\vspace{-0.5em}
\subsection{Why Do One-Shot Update-Vectors Decompose into Task and Domain-Specific Directions}
\label{appx:why_decompose}

Intuitively, VLA inputs contain distinct but overlapping token subsets that are mainly associated with either \textit{task} information (\eg, language instructions and task-relevant objects) or \textit{domain} information (\eg, background appearance, camera viewpoint, and robot embodiment).
Thus, one-shot fine-tuning can induce partially disentangled weight directions~\cite{ortiz2023tangent}.

\begin{figure}[h!]
  \centering
  \vspace{-0.9em}
  \makebox[\columnwidth][c]{%
  \begin{minipage}[c]{0.43\columnwidth}
      \centering
      \vspace{0.5em}
      \setlength{\tabcolsep}{3pt}
      \begin{tabular}{lcc}
        \hline
        Token & \makecell{Prototype\\added to $\theta_0$} & \makecell{Feature\\$L_2$ dist.} \\
        \hline
        \multirow{2}{*}{Text} 
          & \texttt{Domain} & 27.98  \\
          & \texttt{Task}   & \textbf{30.24}  \\
        \hline
        \multirow{2}{*}{Image} 
          & \texttt{Domain} & \textbf{28.03}  \\
          & \texttt{Task}   & 27.11  \\
        \hline
      \end{tabular}
      \captionof{table}{\textbf{Average feature shifts after adding prototypes to $\theta_0$.}}
      \label{tab:feature_dist}
    \end{minipage}
    \hspace{0.01\columnwidth}%
    \begin{minipage}[c]{0.56\columnwidth}
      \centering
      \vspace{-0.4em}
      \includegraphics[width=\linewidth]{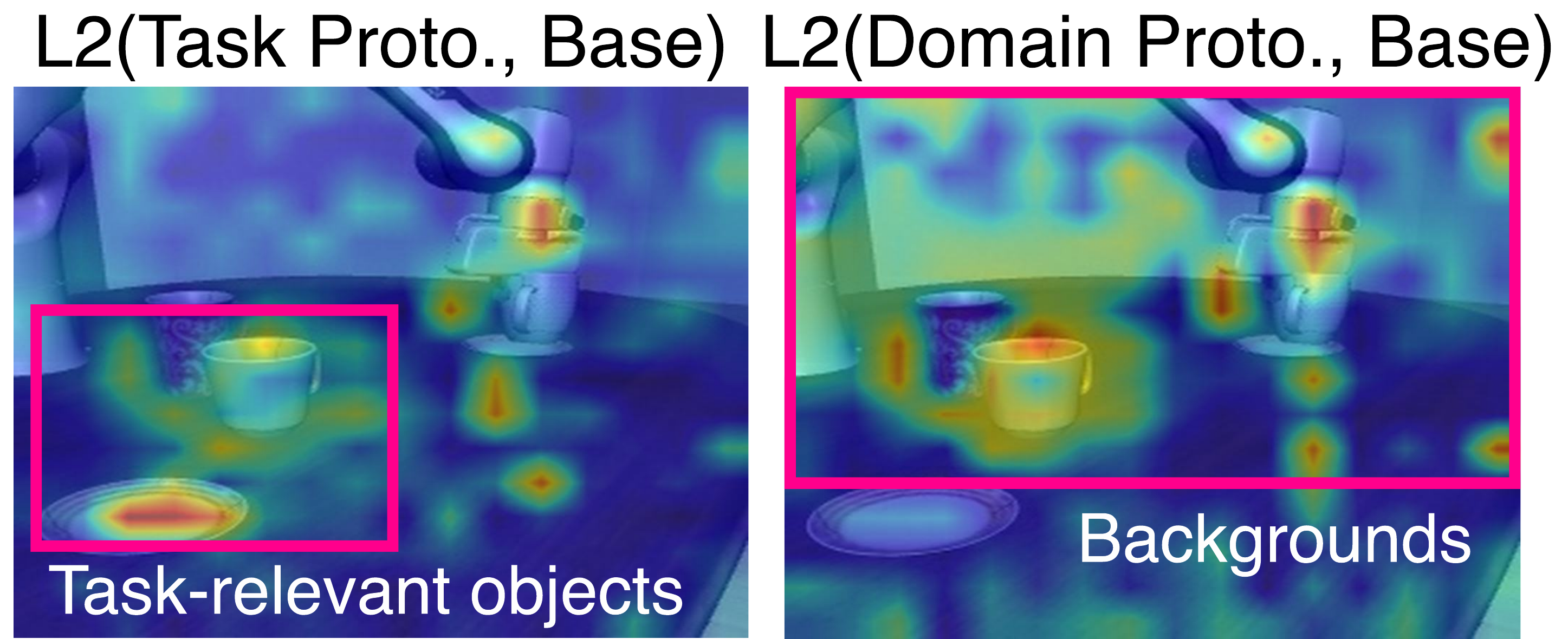}
      \caption{\textbf{Visualization of feature shifts induced by task and domain prototypes.}}
      \label{fig:feature_quali}
    \end{minipage}%
  }
  \vspace{-0.5em}
  \label{fig:coarse}
\end{figure}

To empirically inspect this property, we add each prototype update-vector (Sec.~4.2) to the base model $\theta_0$ and measure last-layer token feature shifts---the $L_2$ distance between features before and after adding the prototype---across different token types. 
As shown in \cref{tab:feature_dist} and \cref{fig:feature_quali}, the \texttt{Task} prototype predominantly shifts text and task-relevant object tokens, while the \texttt{Domain} prototype predominantly shifts background tokens. These results are consistent with recent findings that distinct tasks activate separable column subspaces of weight matrices and that weight interpolation affects intermediate features and functional outputs approximately linearly in pretrained models (based on NTK theory)~\cite{liu2026understanding,zhou2024emergence}, together implying that task- and domain-specific directions can be approximately decomposed and recomposed in weight space.
Furthermore, it motivates DART's core operation of subtracting $\mathrm\Delta_{m,\mathrm{src}}$ from $\mathrm\Delta_{m,\mathrm{tgt}}$ to isolate the target-domain direction $\delta_{\mathrm{tgt}}$.

\vspace{-0.5em}
\section{Details on Baseline Methods}
\label{appx:baselines}

\vspace{-0.5em}
\subsection{RETAIN~\cite{yadav2025retain}}
\label{appx:baseline_retain}

\vspace{-0.5em}
RETAIN is a parameter merging method for VLA models that enables learning a new task while mitigating forgetting of previously learned tasks. It interpolates between the original model parameters and the fine-tuned parameters, balancing knowledge acquisition from the fine-tuned model with retention of the original multi-task capability.

In our experiments, we treat a target-domain task as the new task. Specifically, we first obtain a one-shot fine-tuned model using the same setting as ours, and then merge its parameters with the original source model following RETAIN.

For the scaling coefficient $\alpha$, we independently sweep the coefficients applied to the vision encoder, LLM, and action expert modules, and report the best-performing combination. The selected coefficients are $0.6$, $0.4$, and $0.2$ for the vision, LLM, and action expert modules, respectively.

\vspace{-0.5em}
\subsection{FLA~\cite{li2025fla}}
\label{appx:baseline_fla}

\vspace{-0.5em}
FLA is a parameter-efficient method for adapting VLA models to new domains. 
It achieves strong performance by inserting LoRA~\cite{hu2022lora} layers into the vision encoder while freezing the remaining model parameters.
It is originally designed to adapt VLA models to a new environment using task-wise one-shot demonstrations.

Since we consider a more restrictive data-limited setting, we apply FLA under the same scene-wise one-shot protocol as our method. 
Specifically, we use only one demonstration per scene (i.e., the total number of demonstrations equals the number of scenes), rather than one demonstration per task as originally done in FLA. This ensures a fair comparison under an identical data budget.

We reproduce FLA following the implementation details described in the paper~\cite{li2025fla}. 
To verify correctness, we evaluate our implementation under the experimental protocol reported in the original paper (\ie, using task-wise one-shot demonstrations). 
Our reproduced model achieves performance consistent with the reported results, as summarized in Table~\ref{tab:fla_verification}.

\begin{table}[t!]
\centering
\caption{\textbf{Verification of FLA implementation.}
Comparison between the success rates (\%) reported in \cite{li2025fla} and our reproduced results under the same experimental setup.}
\vspace{-0.5em}
\label{tab:fla_verification}
\setlength{\tabcolsep}{8pt}
\begin{tabular}{lcccc}
\toprule
 & \multicolumn{4}{c}{Novel Viewpoints (Success Rate, \%)} \\
\cmidrule(lr){2-5}
Method (\piOfive) & \texttt{Small} & \texttt{Medium} & \texttt{Large} & Average \\
\midrule
FLA (Reported)   & 94.6 & 90.0 & 87.9 & 90.8 \\
FLA (Reproduced) & 96.0 & 90.8 & 87.9 & 91.6 \\
\bottomrule
\end{tabular}
\end{table}

\vspace{-0.5em}
\section{Experiment Setup Details}
\label{appx:additional_setup}

\vspace{-0.5em}
\subsection{VLA Model and Training Hyperparameter Details}
\label{appx:hyperparam}

\vspace{-0.5em}
We use two VLA models, \piOfive~\cite{zhou2025pi05} and \pifast~\cite{pertsch2025fast}.
All training and evaluation are conducted using the official \texttt{openpi} codebase\footnote{\url{https://github.com/Physical-Intelligence/openpi}}, implemented in JAX.
We use the default model architectures without architectural modifications.
All models take two RGB images (third-person and wrist views) and a language instruction as input, and output an action chunk of 7D vectors $(\mathrm{\Delta}x, \mathrm{\Delta}y, \mathrm{\Delta}z, \mathrm{\Delta}\mathrm{roll}, \mathrm{\Delta}\mathrm{pitch}, \mathrm{\Delta}\mathrm{yaw}, g)$, where $g$ denotes the gripper command.

\begin{table}[t!]
  \centering
  \small
  \caption{\textbf{Training hyperparameters across setups.}
  We use AdamW~\cite{loshchilov2017adamw} with batch size $64$.
  One-shot fine-tuning covers LIBERO~\cite{liu2023libero}, MimicGen~\cite{mandlekar2023mimicgen}, and Real-world.
  Image resolution/action horizon: LIBERO use $256\times 256/10$, MimicGen use $224\times 224$/10, and Real-world uses $224\times 224$/20.
  Real-world uses a decay LR of $2.5{\times}10^{-6}$.}
  \label{tab:hparams_all}
  \vspace{-0.6em}
  \setlength{\tabcolsep}{7pt}
  \renewcommand{\arraystretch}{1.10}

  \begin{tabular}{l l c r r}
    \toprule
    \textbf{Setup} & \textbf{Model} & \textbf{Peak LR} & \textbf{Warmup} & \textbf{Steps} \\
    \midrule
    One-shot fine-tuning   & \piOfive, \pifast & $5{\times}10^{-5}$ & 0 & 1{,}000 \\
    LIBERO source training & \pifast & $5{\times}10^{-5}$ & 10{,}000 & 30{,}000 \\
    MimicGen pretraining   & \piOfive & $5{\times}10^{-5}$ & 10{,}000 & 30{,}000 \\
    MimicGen source training & \piOfive & $5{\times}10^{-5}$ & 2{,}000 & 10{,}000 \\
    Real-world training    & \piOfive & $2.5{\times}10^{-5}$ & 1{,}000 & 10{,}000 \\
    \bottomrule
  \end{tabular}

  \vspace{-0.4em}
\end{table}

We use AdamW~\cite{loshchilov2017adamw} with batch size $64$ for all training runs.
Training hyperparameters are summarized in~\cref{tab:hparams_all}; unless specified in the table, we follow the default settings of the \texttt{openpi} repository.

\vspace{-0.5em}
\subsection{LIBERO Setup Details}
\label{appx:libero_setup}

\vspace{-0.5em}
This section describes the detailed experimental setup on LIBERO~\cite{liu2023libero} for novel viewpoint shifts and combined visual perturbation experiments.
We use the official LIBERO codebase\footnote{\url{https://github.com/Lifelong-Robot-Learning/LIBERO}} with minor modifications to implement domain shifts.

\begin{figure*}[t]
    \centering
    \includegraphics[width=\linewidth]{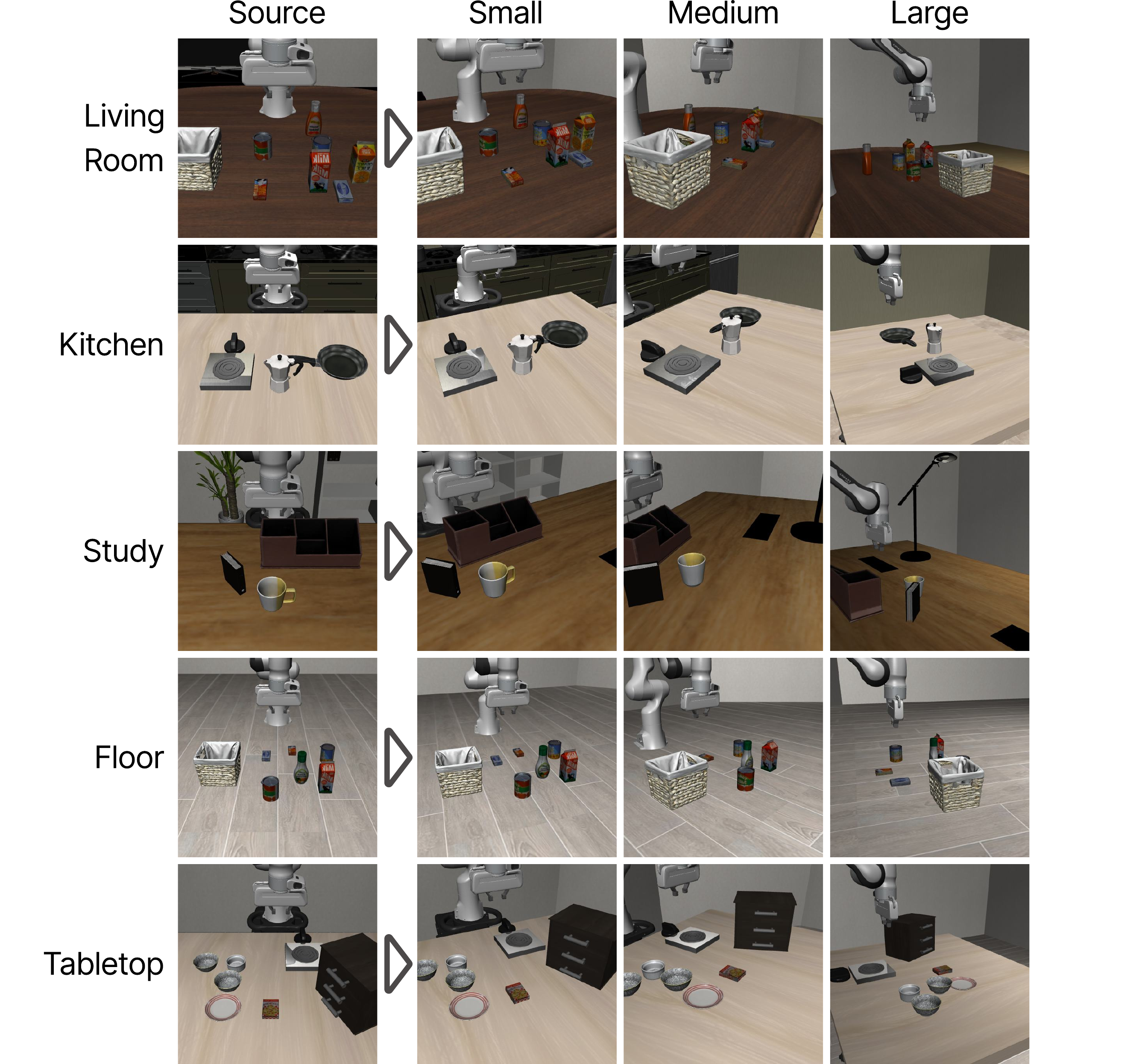}
    \caption{\textbf{Images of each LIBERO scene under different viewpoints.} Columns represent viewpoint shift levels, and rows correspond to scenes.}
    \label{fig:libero_view_preview}
\end{figure*}

\vspace{-0.5em}
\subsubsection{Training and datasets.}
The LIBERO dataset comprises four task suites (\texttt{Spatial}, \texttt{Object}, \texttt{Goal}, \texttt{Long}), each containing 10 tasks with 50 demonstrations per task (2{,}000 demonstrations in total).
These tasks are distributed across five scenes: Living Room, Kitchen, Floor, Study, and Tabletop (\cref{fig:libero_view_preview}).
Specifically, the \texttt{Spatial} and \texttt{Goal} suites are set in the Tabletop scene, \texttt{Object} uses the Floor scene, and \texttt{Long} spans the Living Room, Kitchen, and Study scenes.
We use the filtered version of LIBERO dataset\footnote{\url{https://huggingface.co/datasets/openvla/modified_libero_rlds}{}}, as in OpenVLA~\cite{kim2024openvla}.

As our multi-task source \piOfive model, we adopt the checkpoint trained on the four LIBERO task suites available in \texttt{openpi}\footnote{\url{gs://openpi-assets/checkpoints/pi05_libero}{}}.
For action normalization, we use the statistics provided with the checkpoint and keep them fixed for all subsequent fine-tuning and evaluation.
Since no official \pifast checkpoint pretrained on LIBERO is available, we train the \pifast source model from \texttt{pi0\_fast\_base}\footnote{\url{gs://openpi-assets/checkpoints/pi0_fast_base}{}} on four NVIDIA H100 GPUs.

For one-shot fine-tuning both of \piOfive and \pifast, we use a single demonstration from a single task in each of the five scenes.
As the one-shot demo, we take the first trajectory in each task dataset.
In the target domain, the one-shot fine-tuning uses the regenerated dataset collected under the target domain shift.
We use three different combinations of adaptation tasks for one-shot training and report the average success rate evaluated over all tasks.
The combinations are shown in~\cref{tab:libero_task_comb}.
Fine-tuning is performed on two NVIDIA A100 GPUs.

\vspace{-0.5em}
\subsubsection{Evaluation.}\label{subsubsec:libero_eval}
For each task within a suite, we execute 50 rollout trials using the default initial states provided by the benchmark (10 tasks $\times$ 50 rollouts $=$ 500 rollouts per suite).
An episode is considered successful if the environment returns a done signal (i.e., the task completion condition is met) within the allotted horizon.
The maximum episode horizon is set per suite based on the longest demonstration in the corresponding training set: 220 steps for \texttt{Spatial}, 280 for \texttt{Object}, 300 for \texttt{Goal}, and 520 for \texttt{Long}.
At the start of each episode, we execute 10 no-op steps (zero translation/rotation with gripper open) to allow objects to settle in the simulator before control begins.
We tune the scaling coefficient $\alpha$ on a small set (10 rollouts) in the \texttt{Long} task suite and reuse this value across all tasks in the same setting.
We use $\alpha=0.8$ for viewpoint shifts and $\alpha=0.6$ for visual perturbations.

We report the \emph{Success Rate}, defined as the fraction of trials in which the task is completed, averaged over all tasks within each suite and then averaged across all LIBERO suites.
We use a single NVIDIA A100 GPU for inference and run all experiments in Docker containers to ensure a consistent software environment.

\vspace{-0.5em}
\subsubsection{Implementation of Visual Domain Shifts.}
We modify the camera position and orientation in LIBERO to construct novel viewpoints, following prior work~\cite{wilcox2025adapt3r, li2025fla}. 
Viewpoint shifts are defined as translational offsets of the camera in the MuJoCo~\cite{todorov2012mujoco} simulator. 
Specifically, we define three levels of viewpoint shifts, applied relative to the default camera position:
\begin{itemize}
    \item \texttt{Small}: $(0.0, +0.3, -0.1)$~m
    \item \texttt{Medium}: $(-0.2, +0.7, -0.2)$~m
    \item \texttt{Large}: $(-1.2, +1.0, -0.2)$~m
\end{itemize}
After applying the translation, we rotate the camera to look at the initial end-effector position at $(0.0, 0.0, 0.0)$. 
Because the default camera pose differs across scenes, the resulting rotation angles also vary by scene. 
The full rotation angles are provided in Table~\ref{libero_rotation_angles} and preview images are shown in Figure~\ref{fig:libero_view_preview}.

For \texttt{Light} perturbations, we modify the illumination by increasing the blue-channel intensity in both diffuse and specular components while attenuating the red and green channels. In addition, we reposition the light sources to a single centralized overhead location at a lower height.
We simulate \texttt{Noise} by applying a Gaussian blur ($27\times27$ kernel, $\sigma \approx 4.4$) to all observation images at every timestep, substantially degrading high-frequency visual information such as object edges and textures.

\begin{table}[t!]
\centering
\caption{\textbf{Rotation angles for viewpoint shifts in LIBERO across scenes.} 
Since the default camera pose differs across scenes, the resulting rotation angles vary by scene. 
All values are reported in degrees.}
\label{libero_rotation_angles}
\vspace{-0.5em}
\setlength{\tabcolsep}{8pt}
\begin{tabular}{l ccc}
\toprule
 & \multicolumn{3}{c}{Camera Rotation Angle ($^\circ$)} \\
\cmidrule(lr){2-4}
Scene & \texttt{Small} & \texttt{Medium} & \texttt{Large} \\
\midrule
Living Room & 26.3 & 59.9 & 120.7 \\
Kitchen & 24.5 & 56.8 & 118.4 \\
Study & 33.2 & 69.7 & 126.6 \\
Floor & 18.5 & 45.1 & 106.9 \\
Tabletop & 24.5 & 56.8 & 118.4 \\
\bottomrule
\end{tabular}
\end{table}

\begin{table}[t!]
\centering
\caption{\textbf{Adaptation task combinations in LIBERO used for one-shot fine-tuning.} For LIBERO experiments, we use one demonstration from a single task in each of the five scenes. Success rates are averaged across the three combinations.}
\label{tab:libero_task_comb}
\vspace{-0.5em}
\setlength{\tabcolsep}{8pt}
\begin{tabular}{l p{8cm}}
\toprule
Scene & Task Instruction \\
\midrule

\multicolumn{2}{l}{\small\textbf{Task 1}} \\
\addlinespace[4pt]

Living Room &
\parbox[t][2.2\baselineskip][t]{8cm}{
\texttt{put both the alphabet soup and the tomato sauce in the basket}
}\\
Kitchen &
\parbox[t][2.2\baselineskip][t]{8cm}{
\texttt{turn on the stove and put the moka pot on it}
}\\
Study &
\parbox[t][2.2\baselineskip][t]{8cm}{
\texttt{pick up the book and place it in the back compartment of the caddy}
}\\
Floor &
\parbox[t][2.2\baselineskip][t]{8cm}{
\texttt{pick up the alphabet soup and place it in the basket}
}\\
Tabletop &
\parbox[t][2.2\baselineskip][t]{8cm}{
\texttt{pick up the black bowl between the plate and the ramekin and place it on the plate}
}\\

\midrule
\multicolumn{2}{l}{\small\textbf{Task 2}} \\
\addlinespace[4pt]

Living Room &
\parbox[t][2.2\baselineskip][t]{8cm}{
\texttt{put both the alphabet soup and the cream cheese box in the basket}
}\\
Kitchen &
\parbox[t][2.2\baselineskip][t]{8cm}{
\texttt{put the black bowl in the bottom drawer of the cabinet and close it}
}\\
Study &
\parbox[t][2.2\baselineskip][t]{8cm}{
\texttt{pick up the book and place it in the back compartment of the caddy}
}\\
Floor &
\parbox[t][2.2\baselineskip][t]{8cm}{
\texttt{pick up the ketchup and place it in the basket}
}\\
Tabletop &
\parbox[t][2.2\baselineskip][t]{8cm}{
\texttt{open the middle drawer of the cabinet}
}\\

\midrule
\multicolumn{2}{l}{\small\textbf{Task 3}} \\
\addlinespace[4pt]

Living Room &
\parbox[t][2.2\baselineskip][t]{8cm}{
\texttt{put both the cream cheese box and the butter in the basket}
}\\
Kitchen &
\parbox[t][2.2\baselineskip][t]{8cm}{
\texttt{put the yellow and white mug in the microwave and close it}
}\\
Study &
\parbox[t][2.2\baselineskip][t]{8cm}{
\texttt{pick up the book and place it in the back compartment of the caddy}
}\\
Floor &
\parbox[t][2.2\baselineskip][t]{8cm}{
\texttt{pick up the bbq sauce and place it in the basket}
}\\
Tabletop &
\parbox[t][2.2\baselineskip][t]{8cm}{
\texttt{pick up the black bowl in the top drawer of the wooden cabinet and place it on the plate}
}\\

\bottomrule
\end{tabular}
\end{table}

\vspace{-0.5em}
\subsection{MimicGen Setup Details}
\label{appx:mimicgen_setup}

\begin{figure*}[t!]
    \centering
    \includegraphics[width=\linewidth]{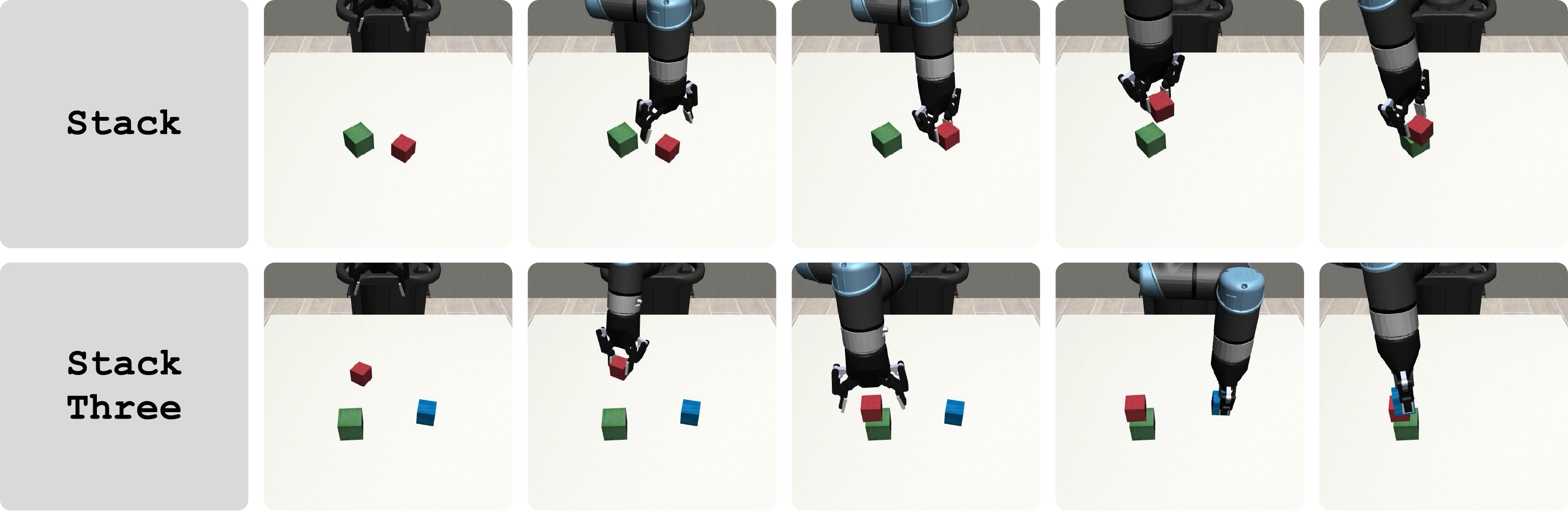}
    \caption{\textbf{Example rollouts on MimicGen.}
    \textbf{Top:} Example rollout of the \texttt{Stack} task with UR5e.
    \textbf{Bottom:} Example rollout of the \texttt{Stack Three} task with UR5e.}
    \label{fig:mimicgen_vis}
\end{figure*}

\vspace{-0.5em}
This section provides dataset, training, and evaluation details for cross-embodiment transfer on MimicGen~\cite{mandlekar2023mimicgen}.
We use the official MimicGen codebase\footnote{\url{https://github.com/NVlabs/mimicgen}} for all experiments.
\Cref{fig:mimicgen_vis} shows example rollouts of the evaluation tasks.

\vspace{-0.5em}
\subsubsection{Training and datasets.}~\label{subsubsec:mimicgen_training}
We pretrain \piOfive on two MimicGen tasks (\texttt{Square} and \texttt{Threading}) on both embodiments under the \texttt{D0} object randomization setting to obtain a stable initialization for our cross-embodiment experiments.
We use 3{,}800 demonstrations for this pretraining stage, all generated using the official MimicGen data generation pipeline.

We then train the source model $\theta_{\text{src}}$ on the Panda robot using the \texttt{Stack} and \texttt{Stack Three} datasets under the \texttt{D0} object initialization (1{,}900 demonstrations in total).
One-shot fine-tuning is conducted on the \texttt{Stack} task using a single demonstration from the Panda dataset and a single demonstration from the UR5e dataset; in both cases, we use the first generated trajectory.

We compute action normalization statistics from the \texttt{Square}/\texttt{Threading} pretraining data and keep them fixed for all subsequent training and evaluation.
All models are trained on two NVIDIA A100 GPUs.

\begin{table}[t!]
\centering
\caption{\textbf{Progress rate rubric on MimicGen.} Progress is the maximum milestone reached within an episode.}
\label{tab:mimicgen_rubric}
\vspace{-0.5em}
\setlength{\tabcolsep}{6pt}
\renewcommand{\arraystretch}{1.15}
\begin{tabular}{l p{0.73\linewidth}}
\toprule
Task & Milestones (Progress., \%) \\
\midrule
\texttt{Stack} &
(50) Grasp the red cube $\rightarrow$ (100) Place it on the green cube. \\
\texttt{Stack Three} &
(25) Grasp the red cube $\rightarrow$ (50) Place it on the green cube $\rightarrow$ \newline
(75) Grasp the blue cube $\rightarrow$ (100) Place it on the red cube. \\
\bottomrule
\end{tabular}
\vspace{-0.5em}
\end{table}

\vspace{-0.5em}
\subsubsection{Evaluation.}
We apply the same action normalization statistics computed in~\cref{subsubsec:mimicgen_training} for all evaluations.
The episode time limit is 200 steps for \texttt{Stack} and 400 steps for \texttt{Stack Three}.
The task descriptions for each task are:
\begin{itemize}
    \item \texttt{Stack}: stack the red cube on the green cube
    \item \texttt{Stack Three}: stack the red cube on the green cube, then stack the blue cube on the red cube
\end{itemize}

We report both \emph{Progress Rate} and \emph{Success Rate}.
For \emph{Progress Rate}, we compute the maximum milestone reached within an episode based on the rubric in~\cref{tab:mimicgen_rubric}, and average it across evaluation rollouts.
As in LIBERO (\cref{subsubsec:libero_eval}), we execute 10 no-op steps to stabilize the environment.
We set $\alpha=0.4$, tuned with 10 rollouts on \texttt{Stack} using a held-out seed.
All experiments are conducted on a single NVIDIA A6000 GPU.

\vspace{-0.5em}
\subsection{Real-world Setup Details}
\label{appx:realworld_setup}

\begin{figure*}[t]
    \centering
    \includegraphics[width=\linewidth]{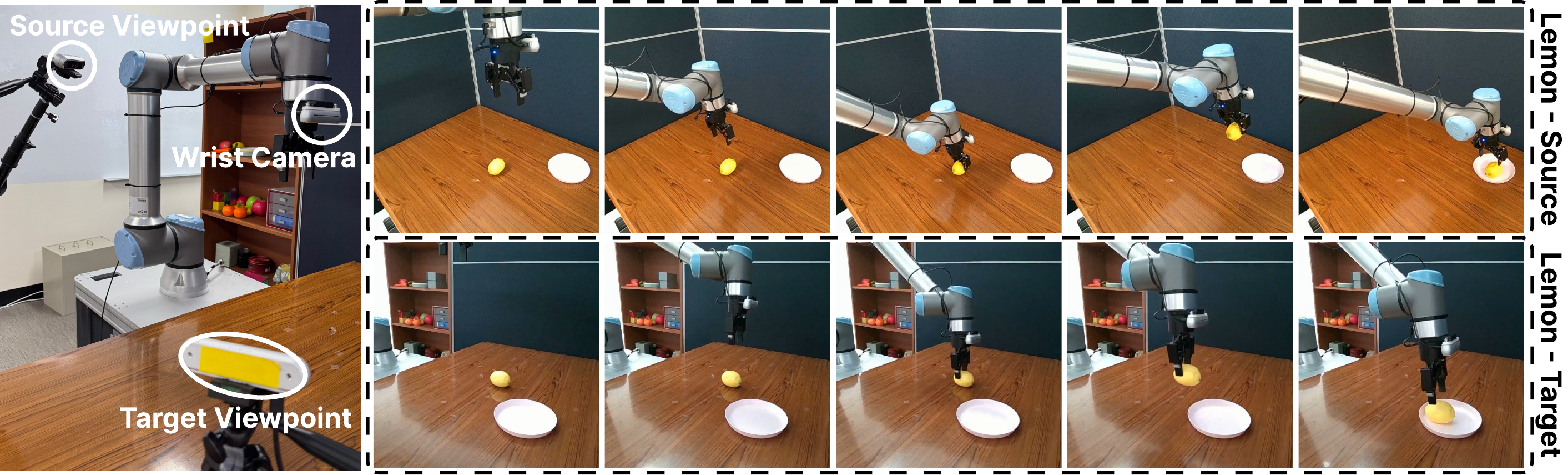}
    \caption{\textbf{Real-world setup and example rollouts.}
    \textbf{Left:} Viewpoint configuration for real-world experiments, using one wrist-mounted camera and a third-person camera (source or target viewpoint).
    \textbf{Right:} Example rollout of the \texttt{Lemon} task from the source viewpoint (top) and the target viewpoint (bottom).}
    \label{fig:realworld_setup}
\end{figure*}

\vspace{-0.5em}
This section outlines details on real-world experiments.
Throughout our experiments, we use a single UR10e arm with a Robotiq 2F-85 gripper.
For vision, we use three RealSense D455 cameras: one wrist-mounted camera and two fixed third-person cameras corresponding to the \texttt{Source} and \texttt{Target} viewpoints, respectively.
\Cref{fig:realworld_setup} shows the viewpoint configuration and example rollouts in both viewpoints.

\vspace{-0.5em}
\subsubsection{Tasks.}
As described in the main paper, we consider three pick-and-place tasks and two fine-grained manipulation tasks.
The pick-and-place task descriptions are:
\begin{itemize}
\vspace{-0.5em}
    \item \texttt{Eggplant}: put the eggplant in the bowl
    \item \texttt{Carrot}: put the carrot on the towel
    \item \texttt{Lemon}: put the lemon on the plate
\end{itemize}
The fine-grained manipulation task descriptions are:
\begin{itemize}
\vspace{-0.5em}
    \item \texttt{Stack Cube}: stack the red cube on the green cube
    \item \texttt{Press Stapler}: press the stapler
\end{itemize}
We use the same fixed language instruction for each task across training and evaluation.
For rollout examples, please see the supplementary video.

\vspace{-0.5em}
\subsubsection{Training and datasets.}
In this experiment, we first train the source policy $\theta_0$ on demonstrations from all five tasks to obtain a multi-task policy, and then perform one-shot fine-tuning for model analogy.
We predefine 12 object positions for each task.
For source-policy training, we collect 120 teleoperated demonstrations (24 per task; 2 demonstrations per position) using Meta Quest 3~\cite{iyer2024openteach}.
For one-shot fine-tuning, we collect a single target-domain demonstration for \texttt{Stack Cube} under the \texttt{Target} viewpoint, and use the corresponding source-domain demonstration from the same predefined position.
We compute action normalization statistics from the source-policy training dataset and reuse them for all subsequent training and evaluation.
Source-model training uses four NVIDIA A100 GPUs, and one-shot fine-tuning uses two NVIDIA A100 GPUs.

\begin{table}[!t]
\centering
\caption{
\textbf{Real-world performance of the base $\pi_{0.5}$ policy on a UR10e robot before adaptation.}
We report success rates (\%) over 12 object positions per task.
\texttt{Source} evaluates the base policy under the \texttt{Source} viewpoint, and \texttt{Target} evaluates the same base policy under the \texttt{Target} viewpoint in a zero-shot manner.
}
\vspace{-0.5em}
\label{tab:realworld_source}

\begin{tabular*}{\linewidth}{@{\extracolsep{\fill}}lcccccc}
\toprule
 & \multicolumn{6}{c}{Success Rate (\%)} \\
\cmidrule(lr){2-7}
 & \multicolumn{3}{c}{Pick-and-Place} & \multicolumn{2}{c}{Fine-grained Manipulation} & \\
\cmidrule(lr){2-4}\cmidrule(lr){5-6}
Viewpoint & \texttt{Eggplant} & \texttt{Lemon} & \texttt{Carrot} & \texttt{Stack Cube} & \texttt{Press Stapler} & Average \\
\midrule
\texttt{Source}  & 100.0 & 91.7 & 100.0 & 100.0 & 100.0 & 98.3 \\
\texttt{Target} (Zero-shot)  & 50.0 & 33.3 & 41.7 & 16.7 & 75.0 & 43.3 \\
\bottomrule
\end{tabular*}
\vspace{-0.5em}
\end{table}

To verify the feasibility of our real-world task setups, we evaluate the base policy $\theta_0$ in the source domain.
As shown in Table~\ref{tab:realworld_source}, the base policy achieves near-perfect success rates on all tasks in the source domain.

\vspace{-0.5em}
\subsubsection{Evaluation.}
We evaluate each policy on the same 12 predefined object positions used during training.
While each model inference predicts the next 20 actions, we execute only the first 15 actions on the real robot.
All experiments are conducted on a single NVIDIA A6000 GPU.

\vspace{-0.5em}
\section{Additional Experiment Results}
\label{appx:additional_analysis}

\subsection{Detailed LIBERO Results by Task Suite}
\label{appx:libero_suitewise}

We report suite-wise success rates on LIBERO (\texttt{Spatial}, \texttt{Object}, \texttt{Goal}, and \texttt{Long}) in 
\cref{tab:viewpoint_task_suites_results}, \cref{tab:visual_shifts_task_suites_results}, and \cref{tab:pifast_viewpoints_task_suites_results}.
Overall, our method outperforms baselines across most task suites and domain-shift settings, indicating consistent improvements.

Interestingly, \pifast (\cref{tab:pifast_viewpoints_task_suites_results}) achieves higher zero-shot success than \piOfive (\cref{tab:viewpoint_task_suites_results}) under viewpoint shifts, aligning with~\cite{fei2025liberoplus}.
Since continuous regression can suffer from compounding feature drift~\cite{lee2024vqbet,shafiullah2022bet,soh2026actionhallucination}, we speculate that \pifast's discrete classification acts as an implicit regularizer, enhancing robustness to visual shifts.

\vspace{-0.5em}
\subsection{Upper Bound Performance of Adaptation}
\label{appx:upperbound}

\begin{table}[t!]
\centering
\caption{
    \textbf{Upper bound performance across experimental settings using \piOfive.}
    We report success rate (\%) on LIBERO and MimicGen for a full fine-tuning upper bound, where the adapted policy $\theta^*$ is obtained by fully fine-tuning the base policy $\theta_0$ using all available target demonstrations.
}
\vspace{-0.6em}
\label{tab:upper_bound}

\setlength{\tabcolsep}{3.7pt}
\renewcommand{\arraystretch}{1.05}

\begin{tabular*}{\linewidth}{@{\extracolsep{\fill}} l l l c c @{}}
\toprule
Setting & Adaptation Type & Task & \# Demos & Success Rate (\%) \\
\midrule

\multirow{8}{*}{LIBERO}
& \multirow{4}{*}{Novel Viewpoint}
& \texttt{Small}  & 1{,}716 & 93.7 \\
& & \texttt{Medium} & 1{,}716 & 91.9 \\
& & \texttt{Large}  & 1{,}716 & 90.9 \\
\cmidrule(l{0pt}){3-5}
& & Average & 1{,}716 & 92.2 \\
\cmidrule(l{0pt}){2-5}

& \multirow{4}{*}{Visual Perturbation}
& \texttt{View}            & 1{,}716 & 91.9 \\
& & \texttt{View+Noise}    & 1{,}716 & 88.3 \\
& & \texttt{View+Noise+Light} & 1{,}716 & 86.1 \\
\cmidrule(l{0pt}){3-5}
& & Average & 1{,}716 & 87.2 \\
\midrule

\multirow{3}{*}{MimicGen}
& \multirow{3}{*}{Cross-Embodiment}
& \texttt{Stack}         & 1{,}900 & 92.4 \\
& & \texttt{Stack Three} & 1{,}900 & 79.6 \\
\cmidrule(l{0pt}){3-5}
& & Average & 1{,}900 & 86.0 \\
\bottomrule
\end{tabular*}

\vspace{-0.6em}
\end{table}
\begin{table}[t!]
\centering
\caption{
    \textbf{Upper bound performance across experimental settings using \pifast.}
    We report success rate (\%) on LIBERO for a full fine-tuning upper bound, where the adapted policy $\theta^*$ is obtained by fully fine-tuning the base policy $\theta_0$ using all available target demonstrations.
}

\label{tab:upper_bound_fast}

\setlength{\tabcolsep}{3.7pt}
\renewcommand{\arraystretch}{1.05}

\begin{tabular*}{\linewidth}{@{\extracolsep{\fill}} l l l c c @{}}
\toprule
Setting & Adaptation Type & Task & \# Demos & Success Rate (\%) \\
\midrule

\multirow{4}{*}{LIBERO}
& \multirow{4}{*}{Novel Viewpoint}
& \texttt{Small}  & 1{,}716 & 87.1 \\
& & \texttt{Medium} & 1{,}716 & 87.9 \\
& & \texttt{Large}  & 1{,}716 & 86.0 \\
\cmidrule(l{0pt}){3-5}
& & Average & 1{,}716 & 87.0 \\
\bottomrule
\end{tabular*}

\vspace{-0.6em}
\end{table}

Although our approach focuses on an one-shot, single-task adaptation, fully fine-tuning on a sufficiently large target-domain dataset is expected to yield the best achievable performance in the target domain.
We use this regime as an empirical \emph{upper bound}.

For each experimental setting, we fully fine-tune the base policy $\theta_0$ on the corresponding target-domain dataset for 10{,}000 steps.
The amount of target-domain data matches the scale used to train the base policy.
For LIBERO~\cite{liu2023libero}, we regenerate the full training set for each target domain using all demonstrations after applying the OpenVLA filtering protocol~\cite{kim2024openvla} (50 demos per task $\times$ 10 tasks $\times$ 4 suites, yielding 1{,}716 demonstrations after filtering).
For MimicGen~\cite{mandlekar2023mimicgen}, we generate 950 target-domain demonstrations per task following the standard MimicGen data generation procedure.

\Cref{tab:upper_bound,tab:upper_bound_fast} reports the resulting success rates and the number of demonstrations used for adaptation.
As expected, full-data fine-tuning consistently yields strong success rate.
However, it requires much more target-domain expert demonstrations, which is impractical at real-world deployment scenarios.
Collecting diverse task-wise demonstrations across many environments is prohibitively expensive, which takes several days to collect and several hours to train.
In contrast, \ours can instantly improve performance in the target domain using only a single demonstration per environment, highlighting its data efficiency.

\vspace{-0.5em}
\subsection{Comparison with Model Merging Methods}
\label{appx:model_merging_methods}

\vspace{-0.5em}
We demonstrate that our proposed model analogy method, \ours, shows superior performance over existing VLA adaptation methods. 
Here, we compare with existing model merging methods, a Task Arithmetic~\cite{ilharco2023TA} branch that merges multiple weights into single weight to combine the capabilities of those weights, to demonstrate that several knowledge interference mitigation approaches~\cite{yadav2023ties, marczak2025isoc, yang2025resm} are not appropriate to extract domain vector.
Specifically, given the update-vectors $\mathrm\Delta_{m,\text{tgt}}$ and $\mathrm\Delta_{m,\text{src}}$, we apply existing model merging methods to combine the two update-vectors $[\mathrm\Delta_{m,\text{tgt}}, -\mathrm\Delta_{m,\text{src}}]$, and we add the combined vector into the base model $\theta_0$ by scaling it with the coefficient $\alpha$. 
We tune the coefficient $\alpha$ in the same way as in our main experiments.

As shown in \cref{tab:merging_methods}, recent state-of-the-art model merging methods show lower success rate compared to our model analogy based method.
This suggests that interference-mitigation strategies for model merging are not directly applicable to model analogy, whose goal is to cancel shared components between source and target updates to isolate the transfer signal.
\begin{table}[t!]
\centering
\caption{\textbf{Comparison with model merging methods under novel viewpoint shifts.}
Average success rates (\%) on LIBERO across three viewpoint shifts (\texttt{Small}, \texttt{Medium}, \texttt{Large}), with the best in \textbf{bold} and the second best \underline{underlined}.}
\vspace{-0.5em}
\label{tab:merging_methods}

\renewcommand{\arraystretch}{1.00}
\setlength{\tabcolsep}{4pt}

\begin{tabularx}{\linewidth}{
>{\raggedright\arraybackslash}p{0.35\linewidth}
*{4}{>{\centering\arraybackslash}X}
}
\toprule
Method (\piOfive)
& \texttt{Small}
& \texttt{Medium}
& \texttt{Large}
& Average \\
\midrule
TIES~\cite{yadav2023ties} \textcolor{blue}{(NeurIPS 2023)} & \underline{91.9} & \underline{79.8} & \underline{61.0} & \underline{77.6} \\
Iso-C~\cite{marczak2025isoc} \textcolor{blue}{(ICML 2025)} & 91.8 & 76.4 & 55.3 & 74.5 \\
RESM~\cite{yang2025resm} \textcolor{blue}{(NeurIPS 2025)} & 90.8 & 78.0 & 57.7 & 75.5 \\
\rowcolor{gray!15}
\ours \textbf{(Ours)} & \textbf{92.0} & \textbf{80.8} & \textbf{64.4} & \textbf{79.1} \\
\bottomrule
\end{tabularx}
\end{table}

\vspace{-0.5em}
\subsection{Comparison with test-time-adaptation method.}
\label{appx:test_time_adaptation}

\vspace{-0.5em}
We compare \ours with SCALE~\cite{choi2026scale} using \pifast~\cite{pertsch2025fast} on LIBERO~\cite{liu2023libero}~(\cref{tab:tta}).
\ours outperforms SCALE across all novel viewpoints, showing the necessity of VLA policy adaptation under environment shifts.

\begin{table}[t!]
\centering
\caption{
\textbf{Comparison with test-time adaptation under novel viewpoint shifts.}
Average success rates (\%) on LIBERO across three viewpoint shifts (\texttt{Small}, \texttt{Medium}, \texttt{Large}) using \pifast~\cite{pertsch2025fast}.
We compare DART with SCALE~\cite{choi2026scale}, with the best in \textbf{bold}.
}
\vspace{-0.5em}
\label{tab:tta}
\setlength{\tabcolsep}{8pt}
\renewcommand{\arraystretch}{1.08}
\begin{tabular}{lcccc}
\toprule
Method (\pifast) & \texttt{Small} & \texttt{Medium} & \texttt{Large} & Average \\
\midrule
SCALE~\textcolor{blue}{(ICML 2026)} & 85.0 & 71.6 & 62.5 & 73.0 \\
\rowcolor{gray!15}
\ours \textbf{(Ours)} & \textbf{91.2} & \textbf{80.8} & \textbf{66.2} & \textbf{79.4} \\
\bottomrule
\end{tabular}
\vspace{-1.0em}
\end{table}

\vspace{-0.5em}
\subsection{Additional Detailed Analysis}
\label{appx:detailed_analysis}

\subsubsection{Source domain performance after adaptation.}
\Cref{tab:forget_mitigate} shows performance on the source domain (i.e., the original environment used for large-scale VLA training) after adaptation with \ours.
\ours performs comparably to the original base policy (Zero-shot) in the source domain.
This suggests that our model analogy framework mitigates forgetting caused by adaptation, indicating that the adapted model can be used across environments, in both source and target domains.

\begin{table*}[t!]
\centering
\small
\caption{
\textbf{Performance on source domain after adaptation in LIBERO using \piOfive.} 
We report the success rate (\%) for each LIBERO task suite on the source domain after adapting the model to each target domain.
Zero-shot refers to the base policy $\theta_0$.
}
\vspace{-0.5em}
\label{tab:forget_mitigate}
\setlength{\tabcolsep}{8pt}
\begin{tabular}{l ccccc}
    \toprule
    & \multicolumn{5}{c}{Novel Viewpoints (Success Rate, \%)} \\
    \cmidrule(lr){2-6}
    Method (\piOfive) & \texttt{Spatial} & \texttt{Object} & \texttt{Goal} & \texttt{Long} & Average \\
    \midrule
    \multicolumn{6}{l}{\textbf{Viewpoint shift:} \texttt{Small}} \\
    \midrule
    Zero-shot                     & 98.8 & 98.2 & 98.0 & 92.4 & 96.9 \\
    \rowcolor{gray!15}
    \ours \textbf{(Ours)}          & 97.6 & 97.2 & 97.0 & 85.8 & 94.4 \\
    \midrule
    \multicolumn{6}{l}{\textbf{Viewpoint shift:} \texttt{Medium}} \\
    \midrule
    Zero-shot                     & 98.8 & 98.2 & 98.0 & 92.4 & 96.9 \\
    \rowcolor{gray!15}
    \ours \textbf{(Ours)}          & 95.8 & 97.6 & 95.6 & 84.4 & 93.4 \\
    \midrule
    \multicolumn{6}{l}{\textbf{Viewpoint shift:} \texttt{Large}} \\
    \midrule
    Zero-shot                     & 98.8 & 98.2 & 98.0 & 92.4 & 96.9 \\
    \rowcolor{gray!15}
    \ours \textbf{(Ours)}          & 98.2 & 98.4 & 95.6 & 83.6 & 94.0 \\
    \bottomrule
\end{tabular}
\vspace{-0.5em}
\end{table*}

\vspace{-0.5em}
\subsubsection{Merging multiple domain vectors with detailed results.}

We investigate whether domain vectors estimated for different target domains can be consolidated into a single vector that generalizes across multiple domains, motivated by our empirical finding that domain directions are relatively disentangled and approximately additive in weight space. 
Specifically, we take three \ours domain vectors $\tilde\delta_{\text{tgt}}$ obtained from LIBERO under novel viewpoint shifts (\texttt{Small}, \texttt{Medium}, \texttt{Large}) and merge them into a combined vector $\delta^{*}$ using recent model-merging techniques~\cite{ilharco2023TA, yadav2023ties, gargiulo2025tsv, marczak2025isoc, yang2025resm}. 
We then adapt the source model $\theta_0$ by adding the merged vector with a scaling coefficient $\alpha$, \ie, $\theta^{*} = \theta_{0} + \alpha \cdot \delta^{*}$.

As shown in \cref{tab:supp_merge_domain_vectors}, the merged domain vector yields moderate and consistent success rates across all three target viewpoint domains, indicating that domain-level updates can be composed into a single transferable direction. 
This further supports the hypothesis that domain directions decompose (approximately) additively.
Practically, it suggests that deployment can maintain one consolidated domain vector for multiple domains, reducing the memory overhead compared to storing a separate vector per target domain.

\begin{table}[t!]
\centering
\caption{\textbf{Performance of merging domain vectors $\tilde\delta_\text{tgt}$ from each novel viewpoint domain into a single adapted model in LIBERO using \piOfive.} 
We report average success rates of total 40 tasks for three trials of one-shot adaptation with different adaptation tasks, with the best in \textbf{bold} and the second best \underline{underlined}.
Merging domain vectors into a single domain vector can adapt the base model to all the target domains. In practice, this property can be used to store only a single composed domain vector across multiple target domains, reducing the memory overhead of storing each domain vector.
}
\vspace{-0.5em}
\label{tab:supp_merge_domain_vectors}
\setlength{\tabcolsep}{8pt}
\begin{tabular}{lcccc}
\toprule
 & \multicolumn{4}{c}{Novel Viewpoints (Success Rate, \%)} \\
\cmidrule(lr){2-5}
Methods (\piOfive) & \texttt{Small} & \texttt{Medium} & \texttt{Large} & Average \\
\midrule
Zero-shot                     & 88.3 & 63.9 & 11.3 & 54.5 \\
\ours (tgt $=$ \texttt{Small})  & \textbf{92.0} & 52.0 & 6.2 & 50.1 \\
\ours (tgt $=$ \texttt{Medium}) & 76.4 & \textbf{80.8} & 10.9 & 56.0 \\
\ours (tgt $=$ \texttt{Large})  & 72.0 & 56.7 & \textbf{64.4} & 64.4 \\
\midrule
\ours & \textbf{92.0} & \textbf{80.8} & \textbf{64.4} & \textbf{79.1} \\
\midrule
\multicolumn{5}{l}{Merging three domain vectors from \ours} \\
\midrule
TA~\cite{ilharco2023TA} \textcolor{blue}{(ICLR 2023)} & 91.3 & 78.7 & 53.5 & 74.5 \\
TIES~\cite{yadav2023ties} \textcolor{blue}{(NeurIPS 2023)} & \underline{91.7} & 75.8 & 46.2 & 70.9 \\
TSV~\cite{gargiulo2025tsv} \textcolor{blue}{(CVPR 2025)} & 90.6 & \underline{79.9} & \underline{56.6} & \underline{75.7} \\
Iso-C~\cite{marczak2025isoc} \textcolor{blue}{(ICML 2025)} & {91.4} & \underline{79.9} & 53.0 & {74.8} \\
RESM~\cite{yang2025resm} \textcolor{blue}{(NeurIPS 2025)} & 91.2 & 76.5 & 40.5 & 69.4 \\
\bottomrule
\end{tabular}
\end{table}

\vspace{-0.5em}
\subsubsection{Different adaptation task from source and target domains.}

So far, we extract the domain vector $\delta_\text{tgt}$ from the target update-vector $\mathrm\Delta_{m,\text{tgt}}$ and the source update-vector $\mathrm\Delta_{m,\text{src}}$ trained on a demonstration of the same adaptation task $\mathcal{T}_m$, \ie, $\mathrm\Delta_{m,\text{tgt}} - \mathrm\Delta_{m,\text{src}}$.
We further study the effect of using a different source update-vector $\mathrm\Delta_{m',\text{src}}$ trained on a demonstration of a different adaptation task $\mathcal{T}_{m'}$, \ie, $\delta_\text{tgt} = \mathrm\Delta_{m,\text{tgt}} - \mathrm\Delta_{m',\text{src}}$. 
To isolate the effect of the source-domain adaptation task, we choose $\mathcal{T}_{m'}$ per scene using two strategies: (i) a \emph{similar-task} choice and (ii) a \emph{random-task} choice.

\vspace{-0.5em}
\paragraph{Setup.}

For similar-task choice, we rank candidate tasks by a feature-level similarity score computed from the last-layer hidden states of the LLM backbone of the VLA model.
Specifically, given a demonstration from task $\mathcal{T}_m$, we extract (i) observation-token features $f^m_o \in \mathbb{R}^{N_o \times d}$ (e.g., image tokens) and (ii) instruction-token features $f^m_I \in \mathbb{R}^{N_I \times d}$ (e.g., language tokens). For two tasks $\mathcal{T}_m$ and $\mathcal{T}_{m'}$, we define:
\begin{equation}\label{eq:cosine_score}
    S(m,m') = \text{avg}\big(\text{cos}(f^m_o, f^{m'}_o) \big) + \text{cos}\big(\text{avg}(f^m_I), \text{avg}(f^{m'}_I) \big)
\end{equation}
where $\text{cos}(\cdot,\cdot)$ denotes cosine similarity. 
For observations, we compute cosine similarity token-wise (since $N_o$ is fixed and token positions align across the image grid) and then average across tokens. 
For instructions, we first average token features into a single vector to handle variable $N_I$.
To reduce noise from single-frame comparisons, we compute $S(m,m')$ at three time steps within each demonstration (first, middle, last frame) and average the resulting scores. Using this score, we select the \emph{most similar} (Top 1) and \emph{second most similar} (Top 2) source-domain tasks to $\mathcal{T}_m$ per scene.

As a contrasting condition, for random-task, we sample $\mathcal{T}_{m'}$ uniformly at random from the source-domain task set per scene.
Together, these two strategies span a controlled range from highly related source tasks to unrelated source tasks, enabling a structured analysis of how the source-task choice affects the resulting domain vector.

\begin{table}[t!]
\centering
\caption{\textbf{Performance of using different adaptation tasks for source and target domain on LIBERO \texttt{Medium} viewpoint using \piOfive.} 
We report average success rates (\%) of total 40 tasks on each target-domain adaptation task $\mathcal{T}_m$ and its corresponding source-domain adaptation task $\mathcal{T}_{m'}$ to extract domain vector $\delta_\text{tgt}  = \mathrm\Delta_{m,\text{tgt}} - \mathrm\Delta_{m',\text{src}}$. $m\in\{1,2,3\}$ is the scene-wise adaptation task combination in \cref{tab:libero_task_comb}. 
For Cosine-Sim, we find the most and the second-most similar tasks per each $m$ using \cref{eq:cosine_score}.
For Random, we randomly choose tasks twice per each $m$.
Despite small task set of LIBERO and simple retrieval heuristic we used, the trend suggests that task similarity is a useful signal for domain vector extraction.
}
\vspace{-0.5em}
\label{tab:src_tgt_combination}
\setlength{\tabcolsep}{8pt}
\begin{tabular}{lcccc}
\toprule
 & \multicolumn{4}{c}{target-domain adaptation task $\mathcal{T}_m$} \\
\cmidrule(lr){2-5}
source-domain adaptation task $\mathcal{T}_{m'}$ & $m=1$ & $m=2$ & $m=3$ & Average \\
\midrule
$m' = m$ & 81.3 & 78.2 & 83.0 & 80.8 \\
$m' = \text{Cosine-Sim: Top 1}$ & 65.3 & 70.1 & 71.6 & 69.0 \\
$m' = \text{Cosine-Sim: Top 2}$ & 69.2 & 66.5 & 67.6 & 67.7 \\
$m' = \text{Random 1}$  & 62.4 & 61.6 & 64.1 & 62.7 \\
$m' = \text{Random 2}$ & 49.3 & 58.1 & 65.8 & 57.7 \\
\bottomrule
\end{tabular}
\end{table}

\vspace{-0.5em}
\paragraph{Results.}
As summarized in \cref{tab:src_tgt_combination}, we evaluate on LIBERO with total 40 tasks using the \piOfive model.
We observe that performance is highest when the source and target adaptation tasks match (\ie, $m'=m$). 
This supports our hypothesis that update-vectors contain largely domain-agnostic task directions, so subtracting $\mathrm\Delta_{m,\text{src}}$ from $\mathrm\Delta_{m,\text{tgt}}$ more effectively cancels task-specific components and isolates the domain vector.
When $m' \neq m$, selecting $\mathcal{T}_{m'}$ via cosine-similarity retrieval consistently outperforms random selection. 
While the gain remains modest—likely due to LIBERO’s limited task set and our simple retrieval heuristic—the trend suggests that task similarity is a useful signal for domain-vector extraction.
Importantly, this result is encouraging for realistic settings where the source-domain data are large and not cleanly indexed by task, making exact matches to the pre-defined adaptation task $\mathcal{T}_m$ from the target-domain unlikely. 
In such cases, retrieving demonstrations of \emph{similar} tasks can still yield meaningful adaptation, and we expect further improvements with stronger retrieval methods~\cite{dass2025datamil,xie2025iwr,kumar2025collage} and with source datasets that contain more diverse tasks (as is typical in large-scale real-world robot datasets~\cite{o2024oxe,walke2023bridgedata,khazatsky2024droid}, which include 527 distinct tasks~\cite{o2024oxe}).

\vspace{-0.5em}
\subsubsection{Impact of Scaling Coefficient Per Domain.}

Beyond the scaling coefficient analysis presented in Fig.~6(a) of the main paper, we provide detailed per-domain performance across varying scaling coefficients $\alpha$. 
As shown in \cref{fig:effect_coef}, while the optimal $\alpha$ differs across domains, performance remains stable over a wide range of values, with standard deviations of 0.8\% and 2.1\% for \texttt{Medium} and \texttt{Large}, respectively.

\vspace{-0.5em}
\subsubsection{Impact of Training Time.}

Beyond the per-training-step analysis in Fig.~6(b) of the main paper, we compare \ours against VLA adaptation baselines under comparable training budgets. Since \ours trains two independent one-shot fine-tuning models—one on source-domain data and one on target-domain data—each model requires fewer than half the training steps of RETAIN and FLA. Despite this reduced per-model training, \ours consistently outperforms both baselines (\cref{fig:effect_traintime}), demonstrating its effectiveness and training efficiency. While this comparison assumes single-GPU training, the total adaptation time can be further reduced by training the source- and target-domain models in parallel when multiple GPUs are available.

\vspace{-0.5em}
\subsubsection{Impact of scaling coefficient and training time.}
\begin{figure}[t!]
  \centering
  \makebox[\columnwidth][c]{%
    \begin{minipage}[c]{0.49\columnwidth}
      \centering
      \includegraphics[width=\linewidth]{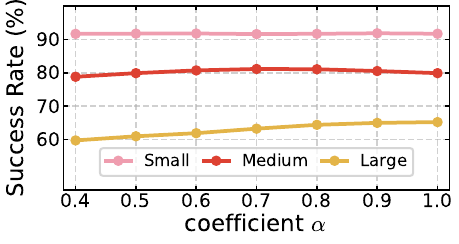}
      \caption{\textbf{Impact of scaling coefficient $\alpha$.}}
      \label{fig:effect_coef}
    \end{minipage}%
    \hspace{0.02\columnwidth}%
    \begin{minipage}[c]{0.49\columnwidth}
      \centering
      \includegraphics[width=\linewidth]{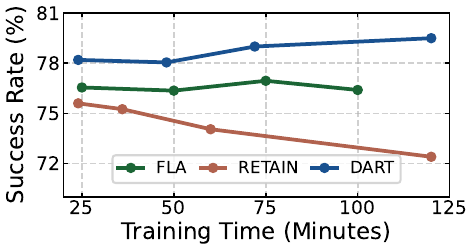}
      \caption{\textbf{Impact of training time.}}
      \label{fig:effect_traintime}
    \end{minipage}%
  }
  \vspace{-0.5em}
\end{figure}

We analyze the robustness of \ours to the scaling coefficient $\alpha$ and training time.
As shown in \cref{fig:effect_coef}, performance remains stable over a wide range of $\alpha$, with standard deviations of $0.8\%$ for \texttt{Medium} and $2.1\%$ for \texttt{Large}.
Under comparable training time, obtained by adjusting training steps, \ours consistently outperforms FLA and RETAIN~(\cref{fig:effect_traintime}).
All methods have the same inference time.

\vspace{-0.5em}
\subsubsection{Choice of layers to adapt.}
\begin{figure}[tb]
    \centering
    \includegraphics[width=\linewidth]{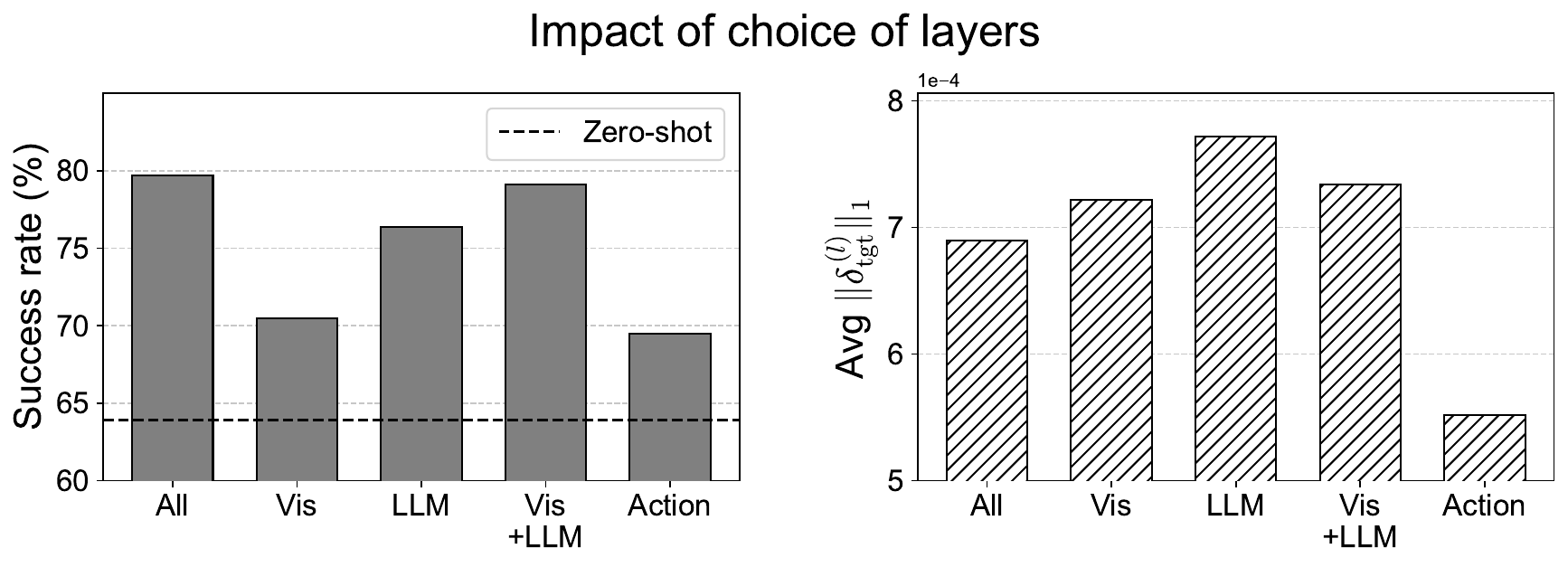}
    \caption{\textbf{Impact of choice of layers to adapt.} 
    \texttt{Vis} is vision encoder, \texttt{LLM} is language model, and \texttt{Action} is action expert in the VLA model, \piOfive. 
    We report the average success rate (\%) on LIBERO in three novel viewpoint shifts (\texttt{Small}, \texttt{Medium}, \texttt{Large}) (Left). We also measure the average absolute value of the domain vectors across the chosen layers (Right).
    }
    \label{fig:effect_layer}
    \vspace{-0.5em}
\end{figure}

We study where to apply the domain vector $\tilde\delta^{(l)}_{\text{tgt}}$ across the vision encoder (\texttt{Vis}), language model (\texttt{LLM}), and action expert (\texttt{Action}) in VLA models. 
As shown in \cref{fig:effect_layer}, updating all layers achieves the best performance, while updating \texttt{Vis+LLM} is nearly identical, indicating that adapting \texttt{Action} provides only marginal benefit. 
This is consistent with the smallest mean absolute magnitude of $\tilde\delta^{(l)}_\text{tgt}$ for \texttt{Action} and its low performance when adapted alone, whereas \texttt{Vis} and especially \texttt{LLM} have larger mean $\|\tilde\delta^{(l)}_{\text{tgt}}\|_1$ and account for most of the gain. 
The modest improvement from \texttt{Vis}-only adaptation suggests that viewpoint shifts affect not only perception but also language-conditioned downstream decisions, emphasizing the role of \texttt{LLM} adaptation.

\vspace{-0.5em}
\subsubsection{Per-layer subspace alignment score $\gamma^{(l)}$.}
\cref{fig:perlayer_sar} shows the subspace alignment score $\gamma^{(l)}(\mathrm\Delta_{m,\text{src}},\mathrm\Delta_{m,\text{tgt}})$ per layer in a VLA model, which we use it for subspace filtering and subspace scaling in \ours. We observe that \texttt{MLP/Up\_proj} (and \texttt{MLP/Gate\_proj}) layer is highly misaligned between source and target domains. Specifically, the alignment is lower in \texttt{LLM} compared to that in \texttt{VIS} and \texttt{ACTION}, suggesting that domain-specific knowledge is captured in \texttt{LLM} part, making \texttt{LLM} part adaptation important as shown in \cref{fig:effect_layer}.

\vspace{-0.5em}
\subsubsection{Per-layer overlap energy $e^{(l)}_j$.}
\cref{fig:perlayer_energy} shows the average overlap energy $e^{(l)}_j = \big\lVert {U^{(l)\top}_{\text{tgt}}}^{}\mathbf{u}^{(l)}_{\text{src},j}\big\rVert_2^2$ from Eq. (4) of the main paper, which measures the alignment of each subspace vector in a source-domain update-vector to the subspace of the corresponding target-domain update-vector.
Similar to the subspace alignment score plot, \texttt{MLP/Up\_proj} (and \texttt{MLP/Gate\_proj}) layer exhibit consistently low average overlap energy, indicating that many subspace vectors in the source-domain update-vector are highly misaligned with those of the target domain.
This can be explained by the functional role of MLP layers in transformers: the factual knowledge memorized by the model is stored in the MLP layers, where \texttt{Up\_proj} generates the keys used to retrieve certain values in \texttt{Down\_proj}~\cite{meng2022rome, meng2023memit, li2024rmu}.
If we apply this prior observation, the subspace vectors in \texttt{Up\_proj} responsible for generating domain-specific keys are likely highly specialized to the source domain (as the VLA model is trained to look up domain-specific values in \texttt{Down\_proj}), and thus exhibit low alignment with the corresponding target-domain subspace.

\vspace{-0.5em}
\subsubsection{Number of cutoff vectors in \ours.}
\begin{figure}[t!]
  \vspace{-0.5em}
  \centering
  \begin{subfigure}{\linewidth}
    \includegraphics[width=0.99\linewidth]{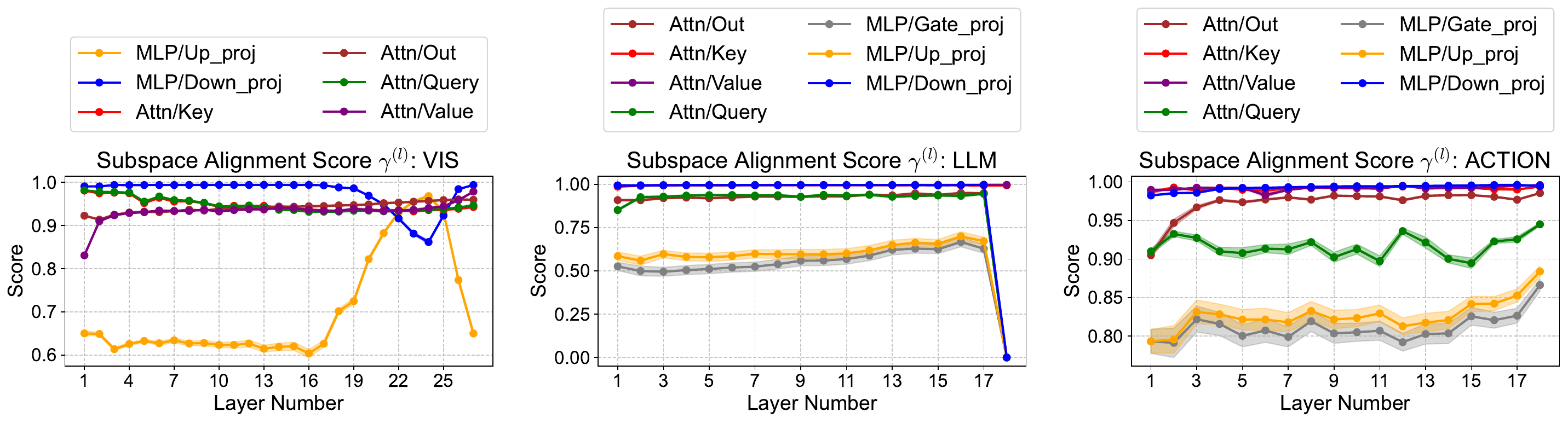}
    \caption{\textbf{Subspace alignment score $\gamma^{(l)}(\theta_{m,\text{src}},\theta_{m,\text{tgt}})$ per layer.}
    }
    \label{fig:perlayer_sar}
  \end{subfigure}
  \hfill
  \vspace{0.25em}
  \begin{subfigure}{\linewidth}
    \includegraphics[width=0.99\linewidth]{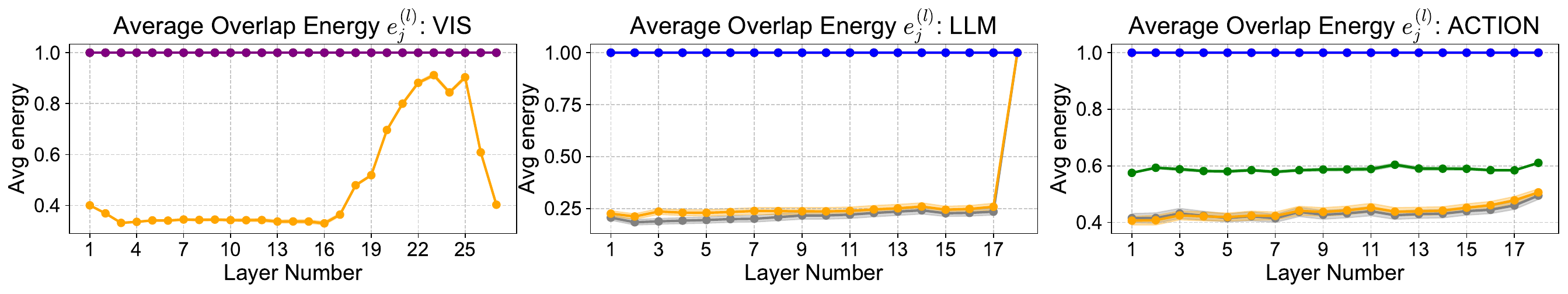}
    \caption{\textbf{Average overlap energy $e^{(l)}_j$ per layer.}
    }
    \label{fig:perlayer_energy}
  \end{subfigure}
  \hfill
  \vspace{0.25em}
  \begin{subfigure}{\linewidth}
    \includegraphics[width=0.99\linewidth]{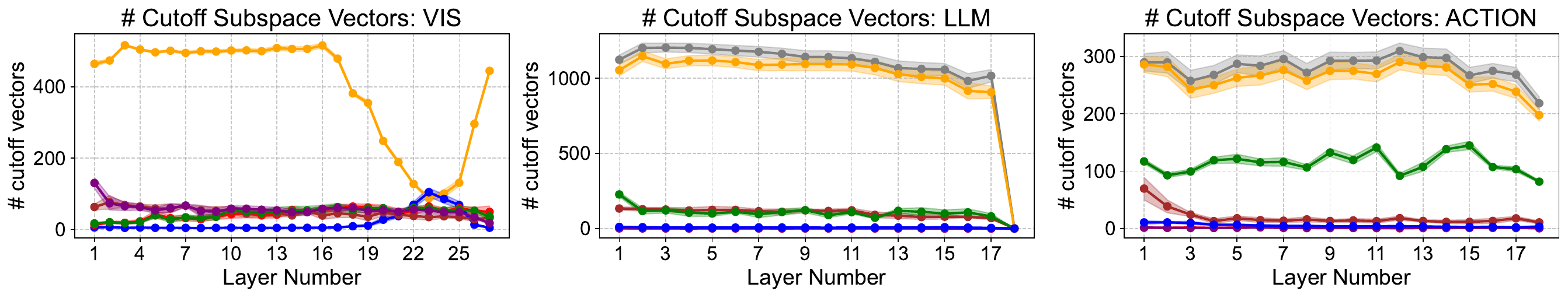}
    \caption{\textbf{Number of cutoff (filtered) subspace vectors per layer.}
    }
    \label{fig:perlayer_cutoff_abs}
  \end{subfigure}
  \hfill
  \vspace{0.25em}
  \begin{subfigure}{\linewidth}
    \includegraphics[width=0.99\linewidth]{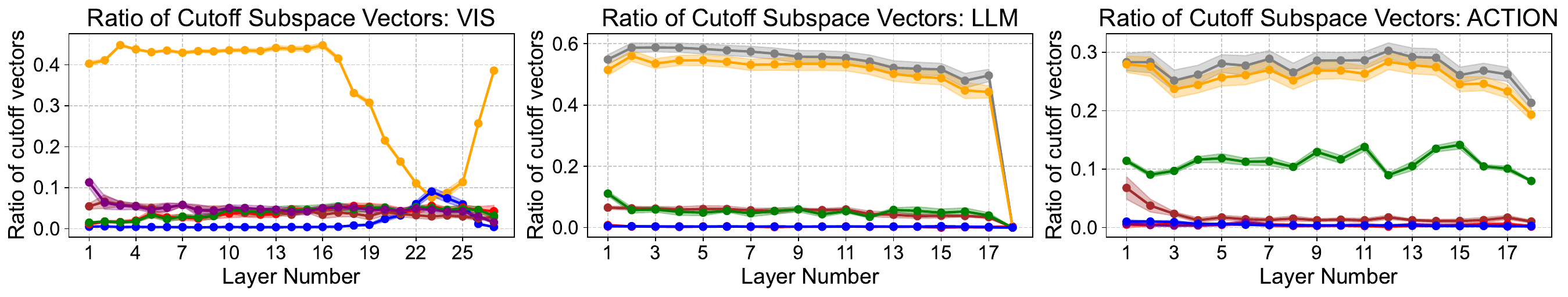}
    \caption{\textbf{Ratio of cutoff (filtered) subspace vectors per layer.}
    }
    \label{fig:perlayer_cutoff_ratio}
  \end{subfigure}
  \caption{\textbf{Per-layer statistics of \ours on LIBERO across novel viewpoints in \piOfive.}
  We plot the mean and standard deviation across three novel viewpoints (\texttt{Small}, \texttt{Medium}, \texttt{Large}) and three different adaptation tasks $\mathcal{T}_m, m\in\{1,2,3\}$.
  \texttt{VIS} is a vision encoder, \texttt{LLM} is a language model, \texttt{ACTION} is an action expert in the VLA model.
  }
  \vspace{-0.5em}
\end{figure}

\cref{fig:perlayer_cutoff_abs} shows the number of cutoff (filtered) subspace vectors per layer in \ours that is decided by the subspace alignment score and the overlap energy. 
As we can naturally expected from the trends observed in \cref{fig:perlayer_sar} and \cref{fig:perlayer_energy}, \texttt{MLP/Up\_proj} (and \texttt{MLP/Gate\_proj}) layer shows that many subspace vectors are filtered before computing the domain vector.
As shown in \cref{fig:perlayer_cutoff_ratio}, the trend remains the same even if we consider the ratio of the removed subspace vectors from the total number of subspace vectors per layer, where some layer types (\eg, \texttt{Attn/Key}, \texttt{Attn/Value}) have smaller dimensions than other types.

\begin{table*}[t!]
\centering
\small
\caption{
\textbf{Suite-wise performance under viewpoint shifts on LIBERO using \piOfive.}
We report success rates (\%) averaged over the tasks in each suite (\texttt{Spatial/\texttt{Object}/\texttt{Goal}/\texttt{Long}}) for each viewpoint (\texttt{Small}/\texttt{Medium}/\texttt{Large}), with the best in \textbf{bold}.
Our method achieves consistent gains across most suites, especially under larger viewpoint shifts.
}
\label{tab:viewpoint_task_suites_results}
\setlength{\tabcolsep}{8pt}
\begin{tabular}{l ccccc}
    \toprule
    & \multicolumn{5}{c}{Novel Viewpoints (Success Rate, \%)} \\
    \cmidrule(lr){2-6}
    Method (\piOfive) & \texttt{Spatial} & \texttt{Object} & \texttt{Goal} & \texttt{Long} & Average \\
    \midrule
    \multicolumn{6}{l}{\textbf{Viewpoint shift:} \texttt{Small}} \\
    \midrule
    Zero-shot                     & 91.2 & 94.4 & \textbf{86.4} & 81.2 & 88.3 \\
    One-shot FT                   & 53.7 & 55.2 & 37.7 & 26.9 & 43.4 \\
    RETAIN~\textcolor{blue}{(ICLR 2026)} & 94.9 & 91.4 & 80.5 & 82.8 & 87.4 \\
    FLA~\textcolor{blue}{(CVPR 2026)}          & 97.7 & 97.1 & 86.0 & \textbf{88.1} & \textbf{92.2} \\
    \rowcolor{gray!15} \ours (\textbf{Ours})    & \textbf{98.4} & \textbf{97.3} & 84.1 & \textbf{88.1} & 92.0 \\
    \midrule
    \multicolumn{6}{l}{\textbf{Viewpoint shift:} \texttt{Medium}} \\
    \midrule
    Zero-shot                     & 65.4 & 87.8 & 69.8 & 32.6 & 63.9 \\
    One-shot FT                   & 39.5 & 54.4 & 23.6 & 15.5 & 33.3 \\
    RETAIN~\textcolor{blue}{(ICLR 2026)} & 78.1 & 92.3 & 63.7 & 55.5 & 72.4 \\
    FLA~\textcolor{blue}{(CVPR 2026)}          & 79.5 & 94.2 & 73.2 & 58.8 & 76.4 \\
    \rowcolor{gray!15} \ours (\textbf{Ours})    & \textbf{87.7} & \textbf{96.3} & \textbf{76.2} & \textbf{62.9} & \textbf{80.8} \\
    \midrule
    \multicolumn{6}{l}{\textbf{Viewpoint shift:} \texttt{Large}} \\
    \midrule
    Zero-shot                     & 10.4 &  0.0 & 28.0 &  6.8 & 11.3 \\
    One-shot FT                   &  13.3 & 34.9 & 11.5 &  11.2 & 17.8 \\
    RETAIN~\textcolor{blue}{(ICLR 2026)} & 41.4 & 73.7 & 36.7 & 43.9 & 48.9 \\  
    FLA~\textcolor{blue}{(CVPR 2026)}          & 52.6 & 69.9 & \textbf{50.7} & 43.9 & 54.3 \\
    \rowcolor{gray!15} \ours (\textbf{Ours})    & \textbf{69.5} & \textbf{87.8} & 46.5 & \textbf{53.9} & \textbf{64.4} \\
    \bottomrule
\end{tabular}
\end{table*}

\begin{table*}[t!]
\centering
\small
\caption{
\textbf{Suite-wise performance under combined visual shifts on LIBERO using \piOfive.}
We report suite-wise success rates (\%) (\texttt{Spatial}/\texttt{Object}/\texttt{Goal}/\texttt{Long}) for each shift setting, with the best in \textbf{bold}.
}
\label{tab:visual_shifts_task_suites_results}
\setlength{\tabcolsep}{8pt}
\begin{tabular}{l ccccc}
    \toprule
    & \multicolumn{5}{c}{Visual Perturbations (Success Rate, \%)} \\
    \cmidrule(lr){2-6}
    Method (\piOfive) & \texttt{Spatial} & \texttt{Object} & \texttt{Goal} & \texttt{Long} & Average \\
    \midrule
    \multicolumn{6}{l}{\textbf{Visual shift:} \texttt{View}} \\
    \midrule
    Zero-shot                     & 65.4 & 87.8 & 69.8 & 32.6 & 63.9 \\
    One-shot FT                   & 39.5 & 54.4 & 23.6 & 15.5 & 33.3 \\
    RETAIN~\textcolor{blue}{(ICLR 2026)} & 78.1 & 92.3 & 63.7 & 55.5 & 72.4 \\
    FLA~\textcolor{blue}{(CVPR 2026)}          & 79.5 & 94.2 & 73.2 & 58.8 & 76.4 \\
    \rowcolor{gray!15} \ours (\textbf{Ours})    & \textbf{87.7} & \textbf{96.3} & \textbf{76.2} & \textbf{62.9} & \textbf{80.8} \\
    \midrule
    \multicolumn{6}{l}{\textbf{Visual shift:} \texttt{View+Noise}} \\
    \midrule
    Zero-shot                     & 70.4 & 88.6 & 51.8 & 30.2 & 60.3 \\
    One-shot FT                   & 44.4 & 42.0 & 14.6 & 9.8 & 27.7 \\
    RETAIN~\textcolor{blue}{(ICLR 2026)} & 82.4 & 81.8 & \textbf{60.6} & \textbf{40.0} & 66.2 \\
    FLA~\textcolor{blue}{(CVPR 2026)}          & 81.8 & \textbf{94.2} & 55.8 & 39.4 & 67.8 \\
    \rowcolor{gray!15} \ours (\textbf{Ours})    & \textbf{90.2} & 93.0 & 55.8 & 37.6 & \textbf{69.2} \\
    \midrule
    \multicolumn{6}{l}{\textbf{Visual shift:} \texttt{View+Noise+Light}} \\
    \midrule
    Zero-shot                     & 65.4 & 82.6 & 69.4 & 11.4 & 57.2 \\
    One-shot FT                   & 44.4 & 42.0 & 14.6 & 13.0 & 28.5 \\
    RETAIN~\textcolor{blue}{(ICLR 2026)} & 76.0 & 92.2 & 61.6 & \textbf{48.2} & 69.5 \\
    FLA~\textcolor{blue}{(CVPR 2026)}          & 80.2 & 92.2 & 71.0 & 37.4 & 70.2 \\
    \rowcolor{gray!15} \ours (\textbf{Ours})    & \textbf{80.8} & \textbf{96.4} & \textbf{74.8} & 48.0 & \textbf{75.0} \\
    \bottomrule
\end{tabular}
\end{table*}

\begin{table*}[t!]
\centering
\small
\caption{
\textbf{Suite-wise performance under combined visual shifts on LIBERO using \pifast.}
We report suite-wise success rates (\%) (\texttt{Spatial}/\texttt{Object}/\texttt{Goal}/\texttt{Long}) for each viewpoint setting, with the best in \textbf{bold}.
}
\label{tab:pifast_viewpoints_task_suites_results}
\setlength{\tabcolsep}{8pt}
\begin{tabular}{l ccccc}
    \toprule
    & \multicolumn{5}{c}{Novel Viewpoints (Success Rate, \%)} \\
    \cmidrule(lr){2-6}
    Method (\pifast) & \texttt{Spatial} & \texttt{Object} & \texttt{Goal} & \texttt{Long} & Average \\
    \midrule
    \multicolumn{6}{l}{\textbf{Viewpoint shift:} \texttt{Small}} \\
    \midrule
    Zero-shot                     & 91.2 & 96.8 & 82.4 & 68.0 & 84.6 \\
    One-shot FT                   & 87.6 & 91.6 & 65.6 & 39.4 & 71.1 \\
    RETAIN~\textcolor{blue}{(ICLR 2026)} & 95.6 & \textbf{98.6} & 82.4 & 76.6 & 88.3 \\
    FLA~\textcolor{blue}{(CVPR 2026)}          & 90.6 & 96.6 & 85.0 & 73.8 & 86.5 \\
    \rowcolor{gray!15} \ours (\textbf{Ours})    & \textbf{96.4} & 98.0 & \textbf{86.2} & \textbf{84.2} & \textbf{91.2} \\
    \midrule
    \multicolumn{6}{l}{\textbf{Viewpoint shift:} \texttt{Medium}} \\
    \midrule
    Zero-shot                     & 78.4 & 95.6 & 79.0 & 41.2 & 73.6 \\
    One-shot FT                   & 75.6 & 94.0 & 52.2 & 30.0 & 63.0 \\
    RETAIN~\textcolor{blue}{(ICLR 2026)} & 79.4 & 97.8 & 81.0 & 55.4 & 78.4 \\
    FLA~\textcolor{blue}{(CVPR 2026)}          & \textbf{83.6} & 95.8 & 81.0 & 53.0 & 78.4 \\
    \rowcolor{gray!15} \ours (\textbf{Ours})    & 80.8 & \textbf{99.0} & \textbf{82.6} & \textbf{60.6} & \textbf{80.8} \\
    \midrule
    \multicolumn{6}{l}{\textbf{Viewpoint shift:} \texttt{Large}} \\
    \midrule
    Zero-shot                     & 46.0 & 91.4 & \textbf{71.0} & 39.6 & 62.0 \\
    One-shot FT                   & 53.6 & 84.4 & 55.8 & 14.8 & 52.2 \\
    RETAIN~\textcolor{blue}{(ICLR 2026)} & 50.4 & 91.2 & 67.4 & 41.6 & 62.7 \\
    FLA~\textcolor{blue}{(CVPR 2026)}          & 52.0 & \textbf{92.0} & 63.8 & \textbf{51.6} & 64.9 \\
    \rowcolor{gray!15} \ours (\textbf{Ours})    & \textbf{61.0} & 90.4 & 70.6 & 42.8 & \textbf{66.2} \\
    \bottomrule
\end{tabular}
\end{table*}

\clearpage

\newpage

\end{document}